\documentclass[journal]{IEEEtran}
\ifCLASSINFOpdf
  \usepackage[pdftex]{graphicx}
\else
\fi
\usepackage{amsmath}

\usepackage{import}
\usepackage{amssymb}
\usepackage{algorithm}
\usepackage{algpseudocode}
\usepackage{bm}
\usepackage{multirow}
\usepackage{xcolor}
\usepackage{scalerel}
\usepackage{pbox}

\makeatletter
    \long\def\@makefntext#1{\parindent 1em\noindent
            \hb@xt@1.8em{%
                \hss\@textsuperscript{\tiny\@thefnmark}}#1}%
\makeatother

\algnewcommand\algorithmicswitch{\textbf{switch}}
\algnewcommand\algorithmiccase{\textbf{case}}
\algdef{SE}[SWITCH]{Switch}{EndSwitch}[1]{\algorithmicswitch\ #1\ \algorithmicdo}{\algorithmicend\ \algorithmicswitch}%
\algdef{SE}[CASE]{Case}{EndCase}[1]{\algorithmiccase\ #1}{\algorithmicend\ \algorithmiccase}%
\algtext*{EndSwitch}%
\algtext*{EndCase}%
\DeclareMathOperator*{\argmax}{argmax}
\DeclareMathOperator*{\argmin}{argmin}

\begin{document}
%
\title{Recovery of Linear Components: Reduced Complexity Autoencoder Designs}
%
%
%

\author{Federico~Zocco and
        Se\'an~McLoone,~\IEEEmembership{Senior Member, IEEE}
\thanks{F. Zocco and S. McLoone are with the Centre for Intelligent Autonomous Manufacturing Systems at Queen's University Belfast, Northern Ireland,
UK. E-mail: \{fzocco01, s.mcloone\}@qub.ac.uk}}

%
%

\markboth{\tiny This work has been submitted to the IEEE for possible publication. Copyright may be transferred without notice, after which this version may no longer be accessible.}%
{Zocco \MakeLowercase{\textit{et al.}}: Recovery of Linear Components}
%



\maketitle

\begin{abstract}
Reducing dimensionality is a key preprocessing step in many data analysis applications to address the negative effects of the curse of dimensionality and collinearity on model performance and computational complexity, to denoise the data or to reduce storage requirements. Moreover, in many applications it is desirable to reduce the input dimensions by choosing a subset of variables that best represents the entire set without any a priori information available. Unsupervised variable selection techniques provide a solution to this second problem. An autoencoder, if properly regularized, can solve both unsupervised dimensionality reduction and variable selection, but the training of large neural networks can be prohibitive in time sensitive applications. We present an approach called Recovery of Linear Components (RLC), which serves as a middle ground between linear and non-linear dimensionality reduction techniques, reducing autoencoder training times while enhancing performance over purely linear techniques. With the aid of synthetic and real world case studies, we show that the RLC, when compared with an autoencoder of similar complexity, shows higher accuracy, similar robustness to overfitting, and faster training times. Additionally, at the cost of a relatively small increase in computational complexity, RLC is shown to outperform the current state-of-the-art for a semiconductor manufacturing wafer measurement site optimization application.            
\end{abstract}


%
\IEEEpeerreviewmaketitle

\section{Introduction}
%
%
%
%


\IEEEPARstart{U}{nsupervised} variable selection and dimensionality reduction are fundamental aspects of data analysis, either as objectives in their own right, for example for sensor selection \cite{ranieri2014near}, factor analysis \cite{harman1976modern}, and data visualisation applications \cite{becker2020robust}, or as preprocessing steps to address the curse of dimensionality and collinearity issues that arise with high dimension data.  Principal Component Analysis (PCA) is a well-established linear technique for unsupervised dimensionality reduction. This employs a linear transformation that maps the original variables in the data onto a new set of orthogonal components that are ordered such that they are successively orientated in the directions of maximum variance in the data \cite{Jolliffe2002}. A reduced dimension model is then obtained by retaining the first $k$ components. Other linear transformation techniques include Fisher's linear discriminant analysis (LDA) \cite{fisher1936use}, canonical correlation analysis (CCA) \cite{hotelling1936relations}, and multidimensional scaling (MDS) \cite{borg2003modern}. For a comprehensive review of linear dimensionality reduction the interested reader is referred to \cite{Cunningham2014}. 

Rather than transforming to a new set of variables, it is often desirable to reduce dimensionality by selecting a subset of the original variables (or features) as this enhances the transparency and interpretability of models \cite{flynn2011max}, and in some applications it is a necessity, for example in sensor selection and measurement plan optimisation problems \cite{krause2008near, susto2019induced}. Since determining the best subset of variables from a candidate set, with respect to a given performance metric, is an NP-hard combinatorial optimisation problem, practical implementations of unsupervised variable selection usually involve approximate solutions such as those provided by greedy search algorithms \cite{Puggini2017, Wei2007, cui2008orthogonal, Whitley2000, zocco2017mean}. 

Linear methods enjoy relatively low computational complexity, but are not able to fully exploit the redundancy in the data when the underlying relationships between variables are non-linear \cite{deebani2017ensemble, nguyen2014multivariate, Wang2017a}. Therefore, nonlinear techniques have been developed to address both dimensionality reduction \cite{Maaten2009, Wang2014a} and variable selection \cite{Gregorova2018, Rosasco2012a, Bach2009, chen2017kernel, yamada2014high, du2015unsupervised, li2016deep}, with neural network based algorithms \cite{hinton2006reducing, masci2011stacked, Han2018, wang2017feature, Sun2017, waleesuksan2016fast} proving particularly popular.

\begin{figure}
\centering
\includegraphics[trim=270 220 440 140, clip, width=0.4\textwidth]{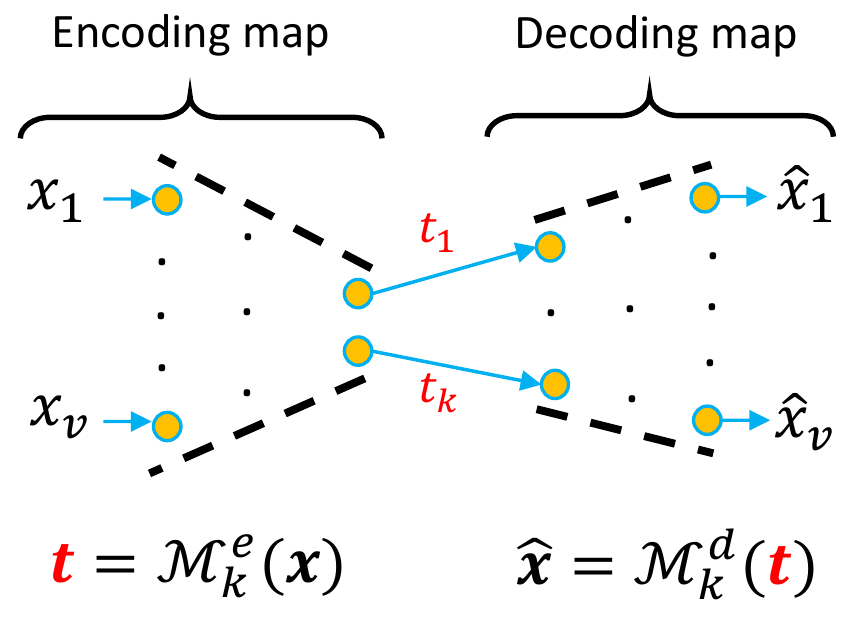}
\caption{Autoencoder neural network topology: the network is trained to reconstruct its own input $\bm{x}$ as its output via a hidden (bottleneck) layer of size $k < v$  whose output  $\bm{t}$ then defines the reduced dimension representation of $\bm{x}$. The encoding and decoding functions, $\mathcal{M}^e_k(\bm{x})$, and $\mathcal{M}^d_k(\bm{t})$, respectively are, in general, nonlinear multilayer neural networks.}
\label{fig:AEscheme}
\end{figure}

The first four of the aforementioned neural network based approaches employ an autoencoder architecture, as depicted in Fig. \ref{fig:AEscheme}. This consists of two sub-networks in cascade: an encoding network, $\bm{t}=\mathcal{M}^e_k(\bm{x})$, that maps a $v$-dimensional input vector $\bm{x}$ to a  $k < v$ dimensional intermediate vector $\bm{t}$; and a decoding network,  $\hat{\bm{x}}=\mathcal{M}^d_k(\bm{t})$, that attempts to reconstruct $\bm{x}$ from $\bm{t}$.
%
%
$\mathcal{M}^e_k(\bm{x})$ and $\mathcal{M}^d_k(\bm{t})$ are in general nonlinear functions whose parameters are determined by training the overall network  to produce the identity mapping, i.e.  $\mathcal{M}^d_k(\mathcal{M}^e_k(\bm{x})) \rightarrow \bm{x}$. Dimensionality reduction is then achieved by virtue of the bottleneck layer of size $k$. 

These neural network based techniques have the advantage of being able to exploit recent advances in deep learning to improve their accuracy, but this comes at the expense of substantially greater computational complexity and training times \cite{Lecun2015, goodfellow2016deep}. Consequently, in this paper we propose an unsupervised dimensionality reduction and feature selection methodology, referred to as Recovery of Linear Components (RLC), that is designed to be a middle ground between autoencoders and linear dimensionality reduction techniques. The methodology involves two stages; first, linear techniques are used to produce an ordered set of $k_{lin}$ linearly transformed components from the original $v$ dimensional dataset; then, the enhanced representational capabilities of neural networks are exploited to predict these  $k_{lin}$ components, from a subset of the first $k$ components. The advantage of this two-stage approach is that it requires less complex networks compared to autoencoders, thereby accelerating the training phase.  

The  main contributions of this paper are the development of the RLC methodology and training algorithm, and the benchmarking of its performance on a range of case studies. In particular:
 \begin{itemize}
\item{we show experimentally that the RLC, when compared with a conventional nonlinear autoencoder of similar complexity, shows higher accuracy, similar robustness to overfitting, and faster training times;}  
\item{for the specific case of a semiconductor manufacturing wafer surface measurement plan optimisation application which motivated this work, we show that RLC outperforms the state-of-the-art, allowing fewer measurements to be taken while still accurately recovering the full wafer profile \cite{Prakash2012a, Puggini2017}}. 
\end{itemize}

  
The remainder of the paper is organized as follows. Section \ref{sec:Prelim} provides a brief description of the dimensionality reduction, variable selection and autoencoder algorithms that underpin the work. The RLC methodology and training algorithm are introduced in Section \ref{sec:PropAlg}. Experimental results are presented in sections \ref{sec:VarSel} and \ref{sec:DimRed} for unsupervised variable selection and unsupervised dimensionality reduction, respectively. Finally, conclusions are provided in Section \ref{sec:Concl}.

\section{Preliminaries}
\label{sec:Prelim}
\noindent The following notation and conventions are adopted throughout the paper. Matrices and vectors are denoted by bold capital and lowercase letters, respectively, while scalars are denoted by non-bold lower case letters. The dimensionality reduction/variable selection problem is with respect to a dataset consisting of  $m$ measurements of $v$ variables collected in a matrix $\bm{X} \in \mathbb{R}^{m \times v}$, hereinafter assumed to be scaled to have zero-mean columns. The objective is to find a lower dimensional representation of $\bm{X}$ that adequately captures the information contained in $\bm{X}$. Denoting this representation as $\bm{T}_k \in \mathbb{R}^{m \times k}$, the encoding process can be expressed as $\bm{T}_k=\mathcal{M}^e_k(\bm{X})$ and the decoding process (i.e. the reconstruction of $\bm{X}$ from $\bm{T}_k$) as $\hat{\bm{X}}=\mathcal{M}^d_k(\bm{T}_k)$.

The adequacy of $\bm{T}_k$ as a representation of $\bm{X}$ can be quantified in terms of the percentage variance explained $V_{EX}(\cdot)$ between the original matrix $\bm{X}$ and its reconstruction $\bm{\hat{X}}$, that is 
\begin{equation}
V_{EX}(\bm{X};\hat{\bm{X}}) = 100 \Big(1 - \frac{||\bm{X}-\hat{\bm{X}}||^2_F}{||\bm{X}||^2_F}\Big),
\end{equation}
where $||\cdot||_F$ denotes the Frobenius norm.  Alternatively the lossiness of $\bm{T}_k$ can be measured in terms of the mean squared reconstruction error $M_{SE}(\cdot)$ defined as 
\begin{equation}
M_{SE}(\bm{X};\hat{\bm{X}}) = \frac{1}{mv}||\bm{X}-\hat{\bm{X}}||^2_F
\end{equation}
This is complementary to $V_{EX}(\cdot)$ since 
\begin{equation}
M_{SE}(\bm{X};\hat{\bm{X}}) = \alpha(100 - V_{EX})\text{, with }\alpha = \frac{||\bm{X}||^2_F}{100mv}, 
\end{equation}
hence, as an optimization objective, maximizing $V_{EX}(\cdot)$ is equivalent to minimizing $M_{SE}(\cdot)$. 

The dimensionality reduction techniques that are the foundation of RLC can be classified as either linear or nonlinear. In the former both the encoding and decoding processes are linear transformations, while in the latter they are nonlinear.   

\subsection{Linear techniques}

\noindent\textbf{Dimensionality reduction}: In linear dimensionality reduction the encoding function is given by
\begin{equation}
\bm{T}_k=\mathcal{M}^e_k(\bm{X})=\bm{X}\bm{P}_k,
\end{equation}
The corresponding decoding transformation is
\begin{equation}
\hat{\bm{X}}=\mathcal{M}^d_k(\bm{T}_k)=\bm{T}_k\bm{B},
\end{equation} 
where
\begin{equation}
\bm{B}=(\bm{T}^\top_k \bm{T}_k)^{-1}\bm{T}^\top_k\bm{X}
\end{equation} 
defines the least squares error linear reconstruction of $\bm{X}$ from $\bm{T}_k$. The encoding transformation matrix $\bm{P}_k$ maximising the variance explained, that is
\begin{equation}
\argmax_{\bm{P}_k} \, V_{EX}(\bm{X}; \hat{\bm{X}}),
\label{Eq:PCAoptimal}
\end{equation}
is then given by the first $k$ principal components (PCs) of the principal component analysis (PCA) of $\bm{X}$ \cite{Jolliffe2002}. These may be computed as the eigenvectors of the sample covariance matrix 
\begin{equation}
\bm{C} = \frac{1}{m-1}\bm{X}^\top \bm{X}
\end{equation}
corresponding to its $k$ largest eigenvalues \cite{smith2002tutorial}. Under these conditions, the decoding transformation reduces to $\bm{B}=\bm{P}^\top_k$. \newline

\noindent \textbf{Variable selection}: When performing dimensionality reduction via variable selection $\bm{T}_k$ is constrained to be a subset of the columns of $\bm{X}$, or equivalently the columns of the encoding transformation matrix $\bm{P}_k$ are constrained to be a subset of the standard basis vectors of $\bm{X}$. A standard basis vector $\bm{e}_i$ is a vector with its $i^\text{th}$ element equal to $1$, and all other elements equal to zero, hence the identity matrix $\bm{I}_v$ can be written as $\bm{I}_v =[\bm{e}_1  \:  \bm{e}_2\:  ... \: \bm{e}_v]$. Using this notation the variable selection optimisation problem can be compactly expressed as 
\begin{equation}
\argmax_{\bm{P}_k \subset \bm{I}_v} \, V_{EX}(\bm{X}; \hat{\bm{X}})
\label{Eq:VSoptimal}
\end{equation}
While finding the optimum subset of $k$ variables from a candidate set of $v$ variables is an NP-hard combinatorial optimisation problem, and therefore computationally intractable in general, satisfactory approximate solutions can be found using greedy search methods such as Forward Selection Component Analysis (FSCA) \cite{Puggini2017}. FSCA performs greedy selection using  $V_{EX}(\cdot)$ as a variable ranking criterion. More precisely, at each iteration it adds to the set of currently selected variables the one that maximizes the variance explained.  Since (\ref{Eq:VSoptimal}) is a constrained version of the PCA optimisation problem (\ref{Eq:PCAoptimal}), it follows that the number of PCs required to achieve a specific value of $V_{EX}(\cdot)$ is a lower bound of the number on variables needed to achieve the same reconstruction accuracy \cite{Puggini2017}. 

To reduce the gap between the FSCA solution and the optimum solution, should one exist, \cite{Puggini2017} propose several refinements to the basic FSCA algorithm. Among these, the two approaches that provided the best accuracy-complexity compromise were the Single-Pass Backward Refinement (SPBR) and the Multi-Pass Backward Refinement (MPBR) algorithms. Since these algorithms can be viewed as refinements of the encoding map $\mathcal{M}^e_k(\bm{X})$ in order to improve its performance, they are complimentary to the methodology  proposed herein as it can be interpreted as a refinement of FSCA that focuses on improving the decoding map $\mathcal{M}^d_k(\bm{T}_k)$.

\subsection{Neural network based nonlinear techniques}
\textbf{Dimensionality reduction}: The autoencoder, as depicted in Fig. \ref{fig:AEscheme}, is a well established approach to achieving unsupervised nonlinear dimensionality reduction. It consists of a multilayer neural network model with a bottleneck layer whose structure and parameters are determined by solving a non-convex optimization problem \cite{hinton2006reducing}. A classical single hidden layer artificial neural network (ANN)  generates a nonlinear mapping between the desired target $\bm{Y} \in \mathbb{R}^{m \times n_y}$ and the input $\bm{X}$ of the form
\begin{equation}
\hat{\bm{Y}}^\top = f_2(\bm{W}_2 f_1(\bm{W}_1\bm{X}^\top + \bm{B}_1) + \bm{B}_2), 
\label{Eq:AEmodel}
\end{equation}      
where, $n_y$ and $n_h$ are the number of neurons in the output and hidden layer, respectively, $\bm{W}_1 \in \mathbb{R}^{n_h \times v}$, $\bm{W}_2 \in \mathbb{R}^{n_y \times n_h}$  are the weight matrices between the input and the hidden layer and hidden layer and the output, respectively, $\bm{B}_1 \in \mathbb{R}^{n_h \times m}$, $\bm{B}_2 \in \mathbb{R}^{n_y \times m}$ are the corresponding bias matrices, $f_l(\cdot)$ is the activation function of the $l$-th layer and $\hat{\bm{Y}}^\top \in \mathbb{R}^{n_y \times m}$ is the network output (i.e. the estimate of the desired target $\bm{Y}$ transposed). When (\ref{Eq:AEmodel}) is configured as a single hidden layer autoencoder,  $\bm{Y} = \bm{X}$ and $n_h = k$. If linear activation functions $f_l(\cdot)$ are used, the autoencoder is linear and yields equivalent $V_{EX}(\cdot)$ performance to PCA. However, the practical value of an autoencoder is its ability to produce more general nonlinear encoding-decoding transformations through the use of nonlinear activation functions such as the logistic sigmoid, the hyperbolic tangent, or the rectified linear unit (ReLU). 

Denoting the output of the autoencoder as
\begin{equation}
\begin{aligned}
\mathcal{N}(\bm{X}; \bm{\theta}) \triangleq [f_2(\bm{W}_2 f_1(\bm{W}_1\bm{X}^\top + \bm{B}_1) + \bm{B}_2)]^\top,
\end{aligned}
\label{Eq:neuralModel}
\end{equation}
where for compactness of notation we introduce
\begin{equation}
\bm{\theta}\triangleq\{\bm{W}_1, \bm{B}_1, \bm{W}_2, \bm{B}_2\},
\end{equation}
the parameters of the optimum autoencoder $\mathcal{N}^*(\bm{X}; \bm{\theta}^*)$ can be computed as 
\begin{equation}
\bm{\theta}^* = \argmin_{\bm{\theta}} J(\bm{X};\bm{\theta})
\label{Eq:optprobAE}
\end{equation}
with $J(\bm{X};\bm{\theta})$ usually chosen as the mean square error cost function $M_{SE}(\bm{X};\hat{\bm{X}}(\bm{\theta}))$, that is
\begin{equation}
J(\bm{X};\bm{\theta}) = \frac{1}{mv}||\bm{X} -\mathcal{N}(\bm{X};\bm{\theta})||^2_F.
\label{Eq:costFcnAE}
\end{equation}
This non-convex optimisation problem is typically solved using a gradient descent algorithm \cite{ruder2016overview}, a quasi-Newton algorithm \cite{bollapragada2018progressive}, or the Levenberg-Marquardt algorithm \cite{more1978levenberg}. 

A single hidden layer autoencoder has limited nonlinear compression capabilities since representational capabilities are a function of the number of hidden neurons, which  is required to be small to achieve dimensionality reduction. To achieve a high level of input space compression, i.e. $k \ll v$, a more complex autoencoder structure has to be designed. Thus, further hidden layers are added to obtain a deep structure called a stacked autoencoder (SAE) \cite{xu2016stacked}. Assuming the general case of an ANN having $L$ layers, the autoencoder equation (\ref{Eq:neuralModel}) becomes 
\begin{equation}
\begin{aligned}
\mathcal{N}(\bm{X}; \bm{\theta}) \triangleq [f_L(\bm{W}_L \dots f_l(\bm{W}_l \dots f_1(\bm{W}_1\bm{X}^\top + \\ + \bm{B}_1) \dots + \bm{B}_l) \dots + \bm{B}_L)]^\top.
\end{aligned}
\end{equation}
However,  increasing the number of layers makes training the autoencoder much more challenging due to the much greater computational complexity, the vanishing gradient problem and the potential to over fit the data when using deep neural networks \cite{goodfellow2016deep}.  Many different strategies have been proposed to address these challenges. Among the most effective are: \emph{early stopping} where training is stopped when the error on a validation set starts to increase \cite{Caruana2001}; introducing different types of operation in the hidden layers such as \emph{convolution} or \emph{pooling} to achieve weight sharing and network size reduction \cite{Luo2018, masci2011stacked}; \emph{Dropout} \cite{srivastava2014dropout} and its extension \emph{DropConnect} \cite{wan2013regularization} where neurons are randomly dropped out with a fixed probability during the training phase as a form of regularisation; and \emph{greedy layer-wise pre-training} \cite{bengio2007greedy, erhan2010does} of networks to improve regularization and training performance through good weight initialization.

Various approaches to achieving greedy layer-wise pre-training have been proposed including using a deep belief net (DBN) \cite{hinton2006fast}, a stacked denoising autoencoder (SDAE) \cite{vincent2010stacked}, a k-sparse autoencoder \cite{Makhzani2013a}, and a sparse autoencoder \cite{Ng2011}. In our work, a sparse autoencoder is used, hence, during pre-training, the basic cost function (\ref{Eq:costFcnAE}) is augmented with regularisation and sparsity terms to become 
\begin{equation}
\begin{aligned}
J(\bm{X};\bm{\theta}) = \frac{1}{mv}||\bm{X} -\mathcal{N}(\bm{X};\bm{\theta})||^2_F + \frac{\lambda}{2}\sum^{L-1}_{l=1}{||\bm{W}_l||^2_F} \\ + \gamma\sum^k_{i=1}{KL(\rho||\hat{\rho}_i)}. 
\end{aligned}
\label{Eq:sparseAE}
\end{equation}
Here, $\lambda$ and $\gamma$ are the weightings applied to the weight decay (i.e. $L_2$ or ridge regression) and sparsity terms, respectively. The sparsity penalty $KL(\rho||\hat{\rho}_i)$, defined as
\begin{equation}
KL(\rho||\hat{\rho}_i) = \rho \text{log}\frac{\rho}{\hat\rho_i} + (1-\rho)\text{log}\frac{1-\rho}{1-\hat{\rho}_i},
\end{equation}     
is the Kullback-Leibler (KL) divergence between two Bernoulli random variables with mean $\rho$ and $\hat{\rho}_i$, respectively, where the random variable mean $\hat{\rho}_i$ is computed as the mean of the output of the $i^\text{th}$ neuron in the bottleneck layer averaged over the training data, and  $\rho$ is the target value for this parameter. This is chosen to be close to zero (e.g. 0.05) to encourage inactivation of the hidden layer neurons, and hence sparsity.

\textbf{Variable selection}: Practically, nonlinear variable selection is not computationally tractable, even using greedy search methods, due to the huge overhead associated with training the nonlinear models for each variable combination evaluated. Consequently, implicit approaches are preferred in the literature when using neural network models, whereby the training cost function $J(\bm{X};\bm{\theta})$ is modified to include a term that induces sparsity in the connection matrix between the inputs and the first hidden layer (i.e. $\bm{W}_1$). For example, in \cite{waleesuksan2016fast,sun2014development} the training phase is split into two phases with the first phase learning the input-output mapping and the second phase focusing on shrinking to zero some of the input layer weights by solving an optimization problem based on the nonnegative garrote parameter. The $L_1$ regularization term, also known as the least absolute shrinkage and selection operator (LASSO), is added to the cost function (\ref{Eq:costFcnAE}) in \cite{Sun2017} to achieve input selection in a Multilayer Perceptron neural network, while \cite{Han2018} and \cite{wang2017feature} integrated a row-sparse regularization term and a graph  regularization trace-ratio criterion, respectively, within the training of an autoencoder to indirectly obtain nonlinear variable selection. Even when using implicit approaches, the computational burden associated with nonlinear variable selection can be prohibitive in time-constrained applications such as dynamic spatial sampling for silicon wafer process monitoring \cite{susto2019induced}, especially when training large neural models. This has motivated our investigation of a methodology that requires the training of less complex nonlinear models, and can be thought as a middle ground between linear and nonlinear variable selection techniques.

\section{Proposed Algorithm}
\label{sec:PropAlg}
\begin{algorithm}
\centering
\caption{Recovery of Linear Components (RLC)}
\label{Alg:RLC}
\begin{algorithmic}[1]
\State \textbf{Input:} $\bm{X}$, $\tau$, $Task$, $k$, neural model hyperparameters 
\State \textbf{Output:} $\bm{P}_{k_{lin}}$, $\mathcal{N}^*$, $\bm\beta$   
\State $k_{lin} = Define\_k_{lin}(\bm{X}, \tau, Task)$ 
\Switch{$Task$}
	\Case{DR}
        \State $\bm{P}_{k_{lin}} = PCA(\bm{X}, k_{lin})$   \hfill $\langle$defines $\bm{P}_{k_{lin}}$$\rangle$   
   \EndCase
	\Case{VS}
       \State $\bm{P}_{k_{lin}} = FSCA(\bm{X}, k_{lin})$   \hfill $\langle$defines $\bm{P}_{k_{lin}}$$\rangle$  
\EndCase
\EndSwitch
\State \textbf{end switch}
\State $\bm{T}_{k_{lin}} = \bm{X}\bm{P}_{k_{lin}}$  \hfill $\langle$uses $\bm{P}_{k_{lin}}$$\rangle$
\If {$k_{lin} \leq k$}    
\Switch{$Task$}
	\Case{DR}
       \State $\bm{\hat{X}} = \bm{T}_{k_{lin}}\bm{P}_{k_{lin}}^\top$  
    \EndCase
	\Case{VS}
           \State $\bm\beta = \bm{T}_{k_{lin}}^+ \bm{X}$ \hfill $\langle$defines $\bm\beta$$\rangle$
           \State $\bm{\hat{X}} = \bm{T}_{k_{lin}} \bm\beta$ \hfill $\langle$uses $\bm\beta$$\rangle$
    \EndCase
\EndSwitch
\State \textbf{end switch}
\Else 
\State $\bm{T}_k$: first $k$ ordered columns of $\bm{T}_{k_{lin}}$
\State $\bm{T}_{\bar{k}}$: remaining $k_{lin} - k$ ordered columns of $\bm{T}_{k_{lin}}$ 
\State Recover the linear space of size $k_{lin}$ by training neural \hphantom{3in} model $\mathcal{N}(\bm{T}_k;\bm\theta)$ to minimize (\ref{Eq:RLCcost})   \hfill $\langle$defines $\mathcal{N}^*$$\rangle$
\State $\bm{\hat{T}}_{\bar{k}} = \mathcal{N}^*(\bm{T}_k;\bm\theta^*)$  \hfill $\langle$uses $\mathcal{N}^*$$\rangle$
\State $\bm\beta = \bm{\hat{T}}_{k_{lin}}^+ \bm{X}$, where $\bm{\hat{T}}_{k_{lin}} = [\bm{T}_k  \,\, \bm{\hat{T}}_{\bar{k}}]$  \hfill $\langle$defines $\bm\beta$$\rangle$
\State $\bm{\hat{X}} = \bm{\hat{T}}_{k_{lin}} \bm\beta$  \hfill $\langle$uses $\bm\beta$$\rangle$
\EndIf   
\end{algorithmic}
\end{algorithm} 
The methodology we propose is called Recovery of Linear Components (RLC). The key idea of our approach is to retain a linear encoding block, but employ a nonlinear model for the decoding block to enhance representational capabilities in the reconstruction phase. Specifically, assuming  that the desired reconstruction accuracy of $\bm{X}$ can be achieved using $k_{lin}$ linear components, then $\bar{k}$ of these are discarded to further reduce the dimensionality to $k < k_{lin}$. Then a nonlinear (neural network) model is used to recover the discarded linear components from the subset that have been retained. In this way, the nonlinear model is only needed to capture the nonlinear relationships between the retained and discarded linear components leading to a much lower complexity model than required for a full $v$ to $v$ autoencoder. In particular, the sum of the input and output dimensions of the RLC neural network model is $k + \bar{k} = k_{lin} < v$, compared to $2v$ for a standard autoencoder. The methodology can be used for both unsupervised dimensionality reduction and unsupervised variable selection if combined with either PCA or FSCA, respectively, to implement the linear encoding step. Hereinafter, the RLC algorithm will be referred to as PCA-RLC or FSCA-RLC whenever it is used in combination with PCA or FSCA, respectively. 

The implementation of RLC is described in Algorithm \ref{Alg:RLC}. The inputs to the algorithm are the data matrix $\bm{X} \in \mathbb{R}^{m \times v}$, the threshold $\tau$ for the desired level of variance explained, the string  $Task \in \{\text{DR}, \text{VS}\}$ indicating whether dimensionality reduction (DR) or variable selection (VS) is being requested,  the number of linear components to be selected $k$, and the neural network hyperparameters (number of layers, number of units per layer, maximum number of training epochs, and regularization parameters).

The algorithm begins by calling the function $Define\_k_{lin}$ which is described in Algorithm \ref{Alg:Defineklin}. This returns $k_{lin}$ as the number of components needed to achieve the desired reconstruction accuracy $\tau$ with the selected linear technique (PCA or FSCA). The goal of RLC is then to achieve this level of accuracy using only $k$ components, where $k < k_{lin}$, allowing $\bar{k} = k_{lin} - k$ components to be discarded. The Nonlinear Iterative Partial Least Squares (NIPALS) algorithm \cite{wold1966estimation}, which computes principal components sequentially in order of significance, can be used to efficiently compute the first $k_{lin}$ principal components if $k_{lin} << v$, otherwise they can be computed using singular value decomposition (SVD) \cite{golub1965calculating} or by the covariance method described in the previous section. 


The outputs of the RLC algorithm are the linear encoding matrix $\bm{P}_{k_{lin}} \in \mathbb{R}^{v \times k_{lin}}$, the linear decoding matrix $\bm\beta$, and the trained neural network model $\mathcal{N}^*$ whose parameters are obtained by minimizing
\begin{equation}
J_{RLC}(\bm{T}_{\bar{k}};\bm\theta) = ||\bm{T}_{\bar{k}} - \mathcal{N}(\bm{T}_k;\bm{\theta})||^2_F.
\label{Eq:RLCcost}
\end{equation}
The matrix $\bm{T}_k \in \mathbb{R}^{m \times k}$ is the reduced dimensionality dataset consisting of either the first $k$ PCs or FSCA selected variables, and $\bm{T}^+_k \in \mathbb{R}^{k \times m}$ is its pseudo-inverse. In the algorithm pseudocode, the text on the right between $\langle \cdot \rangle$ highlights the distinction between the training and test phases. The steps defining the models belong to the training phase whereas steps using the models belong to the test phase.

For benchmarking purposes we implemented two additional algorithms to allow comparisons with more complete deep neural network autoencoder implementations. The first of these, presented in Algorithm \ref{Alg:PCAsae}, tackles the unsupervised dimensionality reduction problem using PCA and a stacked autoencoder (SAE) in cascade, and is called PCA-SAE. PCA reduces the dimension of the input space from $v$ to $k_{lin}$, then the encoder block of the SAE reduces the dimensionality to $k$. The decoder block followed by linear regression are applied to recover $\bm{X}$. 
The training of the SAE is performed in two stages: first a sparse autoencoder based greedy layer-wise pre-training is performed, then the output layer is added and the full network is fine-tuned using the Levenberg-Marquardt training algorithm with early stopping.
 The objective function for the fine-tuning stage is
\begin{equation}
J_{SAE}(\bm{T}_{k_{lin}};\bm{\theta}) = ||\bm{T}_{k_{lin}} - \mathcal{N}(\bm{T}_{k_{lin}};\bm{\theta})||^2_F. 
\label{Eq:costPCA-SAE}
\end{equation}


The second algorithm, Algorithm \ref{Alg:FSCAsde}, tackles the unsupervised variable selection problem. In common with PCA-SAE it employs a deep neural network model and requires a two stage training process, but differs from PCA-SAE by virtue of the fact that the encoding map $\mathcal{M}^e_k(\bm{X})$ is fully linear as in the RLC algorithm, and only the decoder block is nonlinear. The objective function when training the stacked decoder (SDE) block is
\begin{equation}
J_{SDE}(\bm{X};\bm{\theta}) = ||\bm{X} - \mathcal{N}(\bm{T}_{k};\bm{\theta})||^2_F.
\label{Eq:costFSCA-SDE}
\end{equation}
The algorithm, referred to as FSCA-SDE, has been developed to benchmark the RLC algorithm against a technique that shares the same linear encoding map, but exploits the accuracy of a deep neural network in the decoding map $\mathcal{M}^d_k(\bm{T}_{k})$.   

\begin{figure}
\centering
\includegraphics[width=0.45\textwidth]{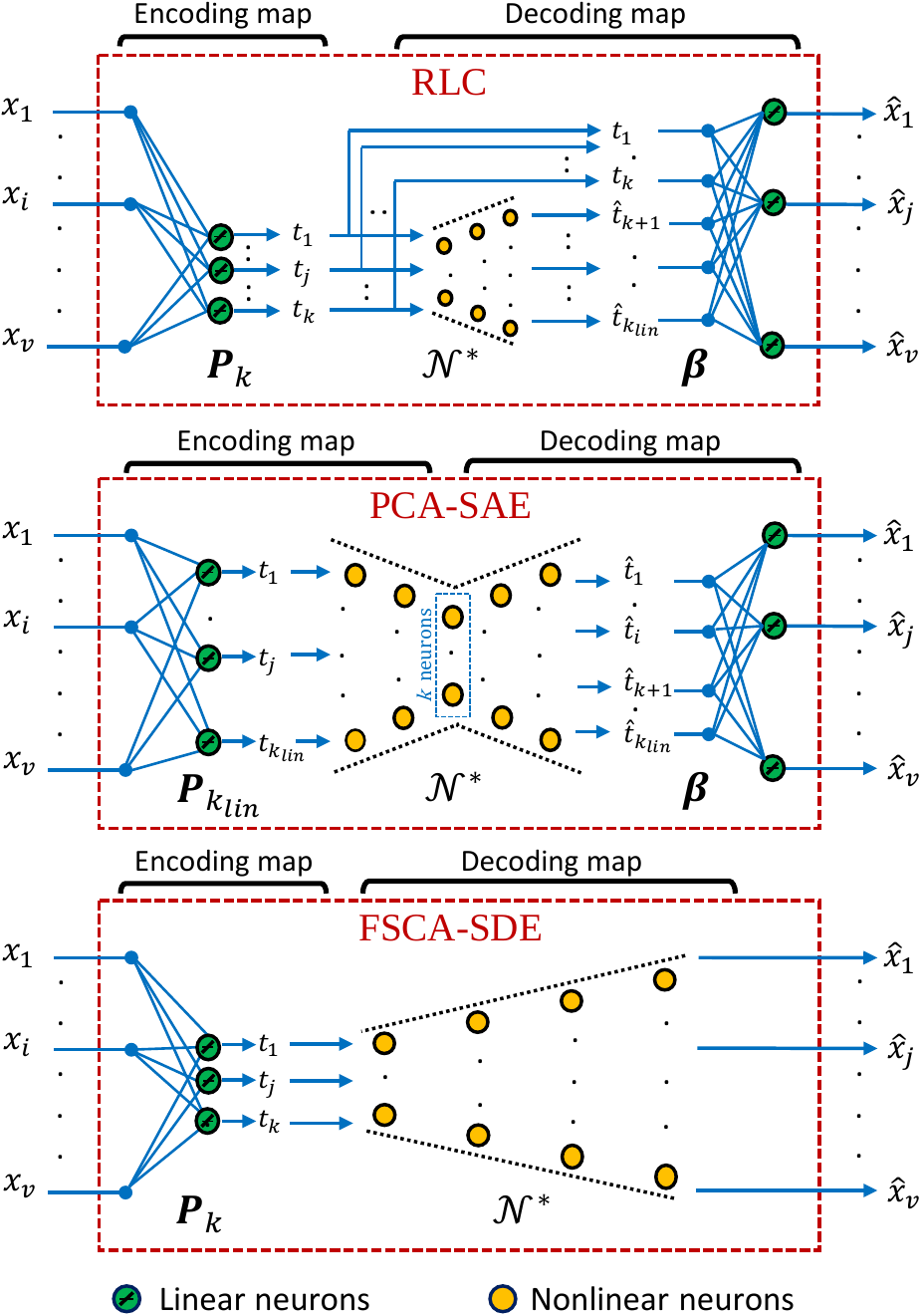}
\caption{Neural network interpretation of RLC, PCA-SAE and FSCA-SDE.} 
\label{fig:AllSchemes}
\end{figure}

The RLC, PCA-SAE and FSCA-SDE architectures can be interpreted as large neural network models consisting of two or three sub-neural models that are independently trained and connected in series. This is illustrated graphically in Fig. \ref{fig:AllSchemes}.

\begin{algorithm}
	\centering
	\caption{$Define\_k_{lin}$}
	\label{Alg:Defineklin}
	\begin{algorithmic}[1]
		\State \textbf{Input:} $\bm{X}$, $\tau$, $Task$ 
		\State \textbf{Output:} $k_{lin}$
		\State $\tilde{k} = 0, \widetilde{V}_{EX} = 0 $
		\While{$\widetilde{V}_{EX} < \tau$}
		\State $\tilde{k} = \tilde{k} + 1$
		\Switch{$Task$}
		\Case{DR}
		\State $\widetilde{V}_{EX} = PCA(\bm{X}, \tilde{k})$ 
		\EndCase
		\Case{VS}
		\State $\widetilde{V}_{EX} = FSCA(\bm{X}, \tilde{k})$
		\EndCase
		\EndSwitch
		\State \textbf{end switch}
		\EndWhile 
		\State $k_{lin} = \tilde{k}$
	\end{algorithmic}
\end{algorithm}

\begin{algorithm}
	\centering
	\caption{PCA-Stacked Autoencoder (PCA-SAE)}
	\label{Alg:PCAsae}
	\begin{algorithmic}[1]
		\State \textbf{Input:} $\bm{X}$, $\tau$, $k$, neural model hyperparameters
		\State \textbf{Output:} $\bm{P}_{k_{lin}}$, $\mathcal{N}^*$, $\bm\beta$ 
		\State $Task$ = `DR'
		\State $k_{lin} = Define\_k_{lin}(\bm{X}, \tau, Task)$ 
		\State $\bm{P}_{k_{lin}} = PCA(\bm{X}, k_{lin})$   \hfill $\langle$defines $\bm{P}_{k_{lin}}$$\rangle$  
		\State $\bm{T}_{k_{lin}} = \bm{X}\bm{P}_{k_{lin}}$   \hfill $\langle$uses $\bm{P}_{k_{lin}}$$\rangle$
		\If {$k_{lin} \leq k$}
		\State $\bm{\hat{X}} = \bm{T}_{k_{lin}}\bm{P}^\top_{k_{lin}}$
		\Else
		\State Greedy layer-wise SAE pre-training via sparse \hphantom{3in}autoencoder (\ref{Eq:sparseAE}) with $\lambda = 0.001, \gamma = 1, \rho = 0.05$ 
		\State Include the output layer and fine-tune the full
		SAE 
		\hspace*{0.15in} $\mathcal{N}(\bm{T}_{k_{lin}};\bm\theta)$ to minimize (\ref{Eq:costPCA-SAE})  \hfill$\langle$defines $\mathcal{N}^*$$\rangle$
		\State $\hat{\bm{T}}_{k_{lin}} = \mathcal{N}^*(\bm{T}_{k_{lin}};\bm\theta^*)$   \hfill $\langle$uses $\mathcal{N}^*$$\rangle$
		\State $\bm\beta = \bm{\hat{T}}_{k_{lin}}^+ \bm{X}$   \hfill $\langle$defines $\bm\beta$$\rangle$
		\State $\bm{\hat{X}} = \bm{\hat{T}}_{k_{lin}} \bm\beta$   \hfill $\langle$uses $\bm\beta$$\rangle$
		\EndIf   
	\end{algorithmic}
\end{algorithm}

\begin{algorithm}
	\centering
	\caption{FSCA-Stacked Decoder (FSCA-SDE)}
	\label{Alg:FSCAsde}
	\begin{algorithmic}[1]
		\State \textbf{Input:} $\bm{X}$, $\tau$, $k$, neural model hyperparameters
		\State \textbf{Output:} $\bm{P}_{k_{lin}}$, $\mathcal{N}^*$, $\bm\beta$ 
		\State $Task$ = `VS'
		\State $k_{lin} = Define\_k_{lin}(\bm{X}, \tau, Task)$ 
		\State $\bm{P}_{k_{lin}} = FSCA(\bm{X}, k_{lin})$ \hfill $\langle$defines $\bm{P}_{k_{lin}}$$\rangle$  
		\State $\bm{T}_{k_{lin}} = \bm{X}\bm{P}_{k_{lin}}$ \hfill $\langle$uses $\bm{P}_{k_{lin}}$$\rangle$
		\If {$k_{lin} \leq k$}
		\State $\bm\beta = \bm{T}_{k_{lin}}^+ \bm{X}$ \hfill $\langle$defines $\bm\beta$$\rangle$
		\State $\bm{\hat{X}} = \bm{T}_{k_{lin}} \bm\beta$ \hfill $\langle$uses $\bm\beta$$\rangle$
		\Else  
		\State $\bm{T}_{k}$: first $k$ ordered columns of $\bm{T}_{k_{lin}}$ 
		\State Greedy layer-wise SDE pre-training via sparse \hphantom{3in}autoencoder (\ref{Eq:sparseAE}) with $\lambda = 0.001, \gamma = 1, \rho = 0.05$  
		\State Include the output decoder and fine-tune the full \hphantom{0in}SDE $\mathcal{N}(\bm{T}_{k};\bm\theta)$ to minimize (\ref{Eq:costFSCA-SDE})  \hfill$\langle$defines $\mathcal{N}^*$$\rangle$
		\State $\hat{\bm{X}} = \mathcal{N}^*(\bm{T}_{k};\bm\theta^*)$   \hfill $\langle$uses $\mathcal{N}^*$$\rangle$
		\EndIf 
	\end{algorithmic}
\end{algorithm}

\section{Results for Unsupervised Variable Selection}
\label{sec:VarSel}
In this section, in addition to FSCA, FSCA-RLC and FSCA-SDE, the SPBR and MPBR algorithms presented in \cite{Puggini2017} are considered for benchmarking purposes. We begin with comparisons on a simulated dataset and then consider 6 real world datasets, 4 of which are publicly available. The algorithms were implemented in MATLAB with MATLAB's Deep Learning Toolbox used to build the autoencoders.  All code developed is available on GitHub\footnote{https://github.com/fedezocco/RecOfLinComp-MATLAB}. The simulations were executed either on an Intel i5-7440HQ CPU with 8 GB of RAM or on an Intel i7-6700 CPU with 16 GB of RAM.
 
\subsection{Synthetic dataset} \label{subsec:var-sel-simulated}
The synthetic dataset, denoted Xsynthetic, is defined as follows. Given generator variables $\{g_i\}_{i=1,2,3} \sim \mathcal{N}(0,1)$ and  noise variables $\{n_i\}_{i=1,2,\dots,10} \sim \mathcal{N}(0,\sigma^2)$, ten nonlinearly related variables are defined as
	
$x_1 = sin(g_1) + n_1$

$x_2 = cos(g_1) + n_2$

$x_3 = x^7_1 + n_3$ 

$x_4 = sign(g_1) \cdot tansig(g_1) + n_4$ 

$x_5 = \frac{\sqrt{|x_3| + |x_1|}}{2} + n_5$

$x_6 = g_1 \cdot x^2_1 \cdot x^3_2 + n_6$ 

$x_7 = cos^3(10g_2) + n_7$ 

$x_8 = \frac{|\frac{g_2}{3125}|-e^{g_2}}{70} + n_8$

$x_9 = \frac{1}{2+|g_1|+e^{g_2}} + n_9$  

$x_{10} = g^3_1 \cdot \frac{g_2}{8} \cdot \frac{g_3}{8} \cdot e^{sign(g_1) \cdot g_3} \cdot cos^2 (g_3) \cdot \frac{sin(g_3)}{44} + n_{10}$

\vspace{1.5mm}
\noindent
Hence, there are three  groups of variables $\{x_1, ..,x_6\}$, $\{x_7,x_8, x_9\}$ and $\{x_{10}\}$, and at least one variable must be selected from each group to enable extraction of the generator variables.
Matrix $\bm{X}^{\text{NL}}  \in \mathbb{R}^{m \times 10}$  is formed from $m$ measurements of these variables. Using a randomly generated linear map $\bm{\Psi} \in \mathbb{R}^{10 \times (v-10)}$, where $\bm{\Psi}_{i,j} \sim \mathcal{N}(0,1)$, and noise matrix $\bm{N} \in \mathbb{R}^{m \times (v-10)}$, where $\bm{N}_{i,j} \sim \mathcal{N}(0,\sigma^2)$, $v-10$ linearly related variables  are generated as $\bm{X}^{\text{L}} = \bm{X}^{\text{NL}}\bm{\Psi} + \bm{N}$. The final dataset is then constituted as $\bm{X} = [\bm{X}^{\text{NL}} \, \, \bm{X}^{\text{L}}] + 70\cdot\bm{1}_m\bm{1}^{\top}_v \in \mathbb{R}^{m \times v}$, where $\bm{1}_{dim}$ denotes a column vector of ones of length $dim$.

\begin{figure}
\centering
\includegraphics[width=0.45\textwidth]{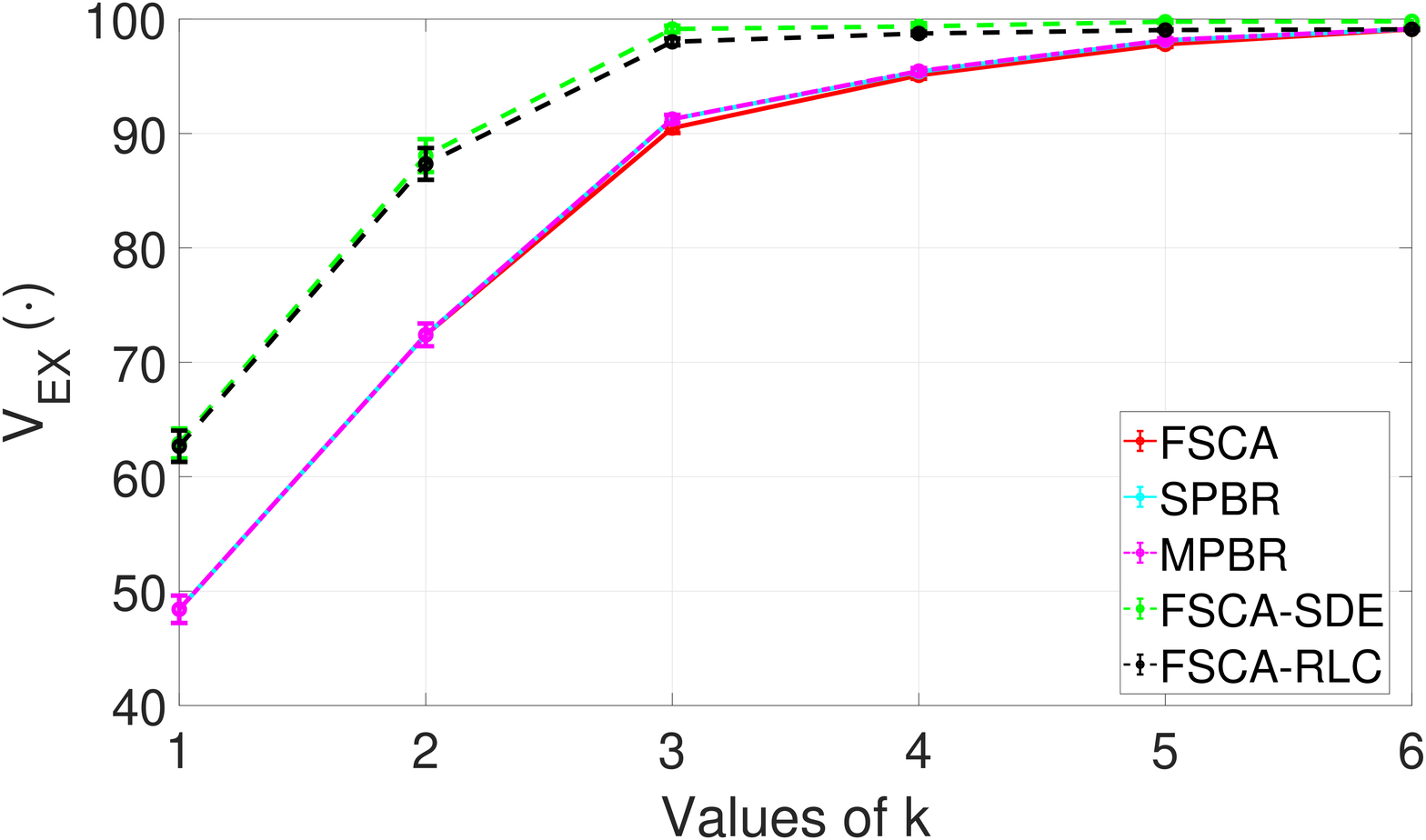}\\
(a) $\sigma^2=0$
\includegraphics[width=0.45\textwidth]{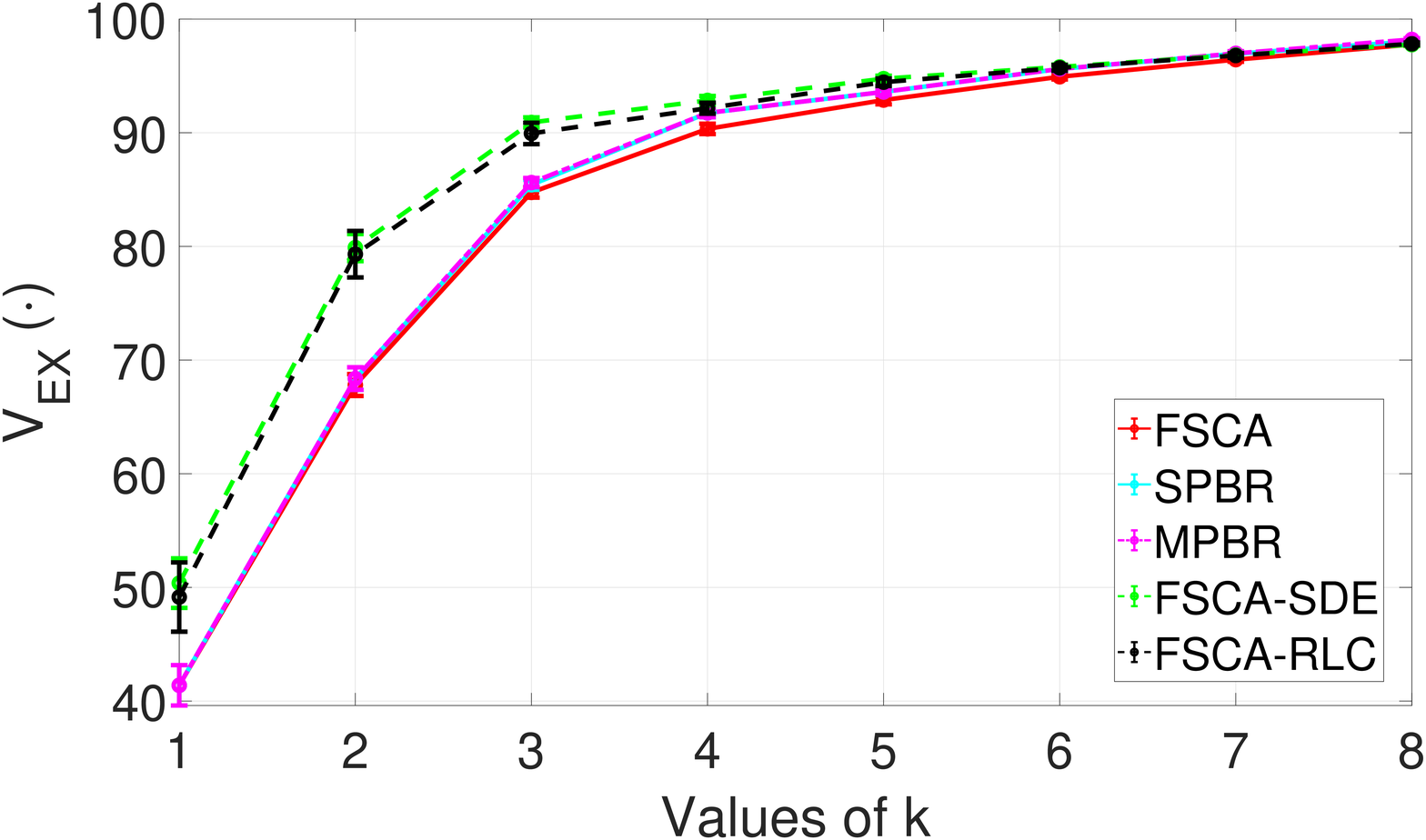}\\
(b) $\sigma^2=0.01$
\caption{Performance of the variable selection algorithms under investigation as a function of $k$ for  Xsynthetic with: (a)  $\sigma^2=0$, and; (b)  $\sigma^2=0.01$. The dots and error bars indicate the mean and standard deviation, respectively, of the variance explained for 100 Monte Carlo simulations.}
\label{fig:VarSel-Simulated}
\end{figure}

\begin{figure}
	\centering
	\includegraphics[width=0.48\textwidth]{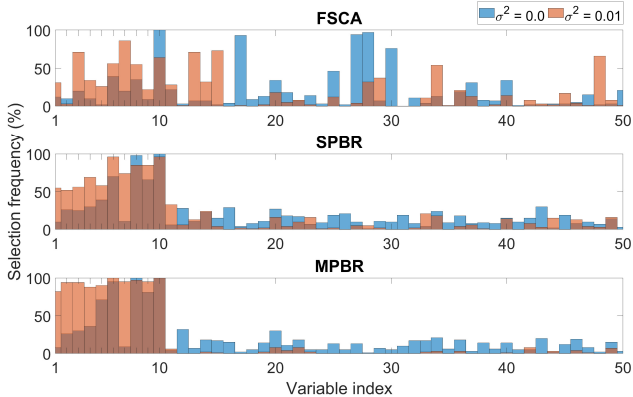}
	\caption{Variable selection frequency (averaged over 100 Monte Carlo simulations) with FSCA, SPBR and MPBR when $k=10$ for  Xsynthetic for $\sigma^2=0$ and $\sigma^2=0.01$. }
	\label{fig:VarSel-Xsimulated1_Histograms}
\end{figure}

The variable selection algorithms have been applied to this dataset with $v$ = 50 and $m$ = 500 for two different noise levels, $\sigma^2$ = 0 and $\sigma^2$ = 0.01, and with the experimental parameters as described in Table \ref{tab:VarSelSettings}. The average and standard deviation (over 100 Monte Carlo simulations) of the variance explained by each algorithm as a function of $k$ is reported in Fig. \ref{fig:VarSel-Simulated}.
FSCA-SDE and FSCA-RLC are the best performing methods with almost equivalent variance explained profiles in terms of mean and standard deviation. SPBR and MPBR perform similarly to FSCA with the exception of $k$ = 3 ($\sigma^2$ = 0) and $k$ = 4 ($\sigma^2$ = 0.01) where they are marginally superior. All methods are relatively robust to variations in the dataset, as suggested by the low standard deviation for each $k$ and their convergence to the same $V_{EX}$ when $k > 5$. In the absence of noise the nonlinear methods are able to achieve  $V_{EX}>98\%$ with 3 variables ($k=3$), whereas 5 variables are required to achieve the same level of accuracy with the linear methods. The frequency with which each variable is selected in the linear encoding step when $k=10$ is shown in Fig. \ref{fig:VarSel-Xsimulated1_Histograms}. This reveals that the first 10 variables are selected with increasing frequency as we move from FSCA to SPBR to MPBR, and are almost exclusively chosen when using MPBR. The focusing of selection on the first 10 variables in more pronounced with the noisy version of the dataset. This is a consequence of the reduced coupling between the first 10 nonlinear variables and the linearly related variables with the addition of noise. While the variables selected by each method vary considerably between methods and between simulation runs they yield similar levels of variance explained, as evident from Fig. \ref{fig:VarSel-Simulated}.

The algorithm computational times, relative to FSCA, are reported in Table \ref{tab:VarSel-OtherDatasets-Time} for several values of $k$ ($\sigma^2=0.01$). FSCA is used as the baseline as it has the lowest computational complexity and is executed as the first step in each of the other methods. SPBR is the quickest method, followed by MPBR, and then FSCA-RLC. FSCA-RLC is 35 times more computationally intensive to compute than MPBR for $k=2$ but becomes more efficient with increasing $k$ and only requires twice the computation time of MPBR for $k = 8$.  In contrast, FSCA-SDE is more than two orders of magnitude more computationally intensive than FSCA-RLC.


\subsection{Real world benchmark datasets}
\begin{figure*}[!t]
\centering
\vspace{5 pt}
\begin{tabular}{ccc}

\includegraphics[width=0.32\textwidth]{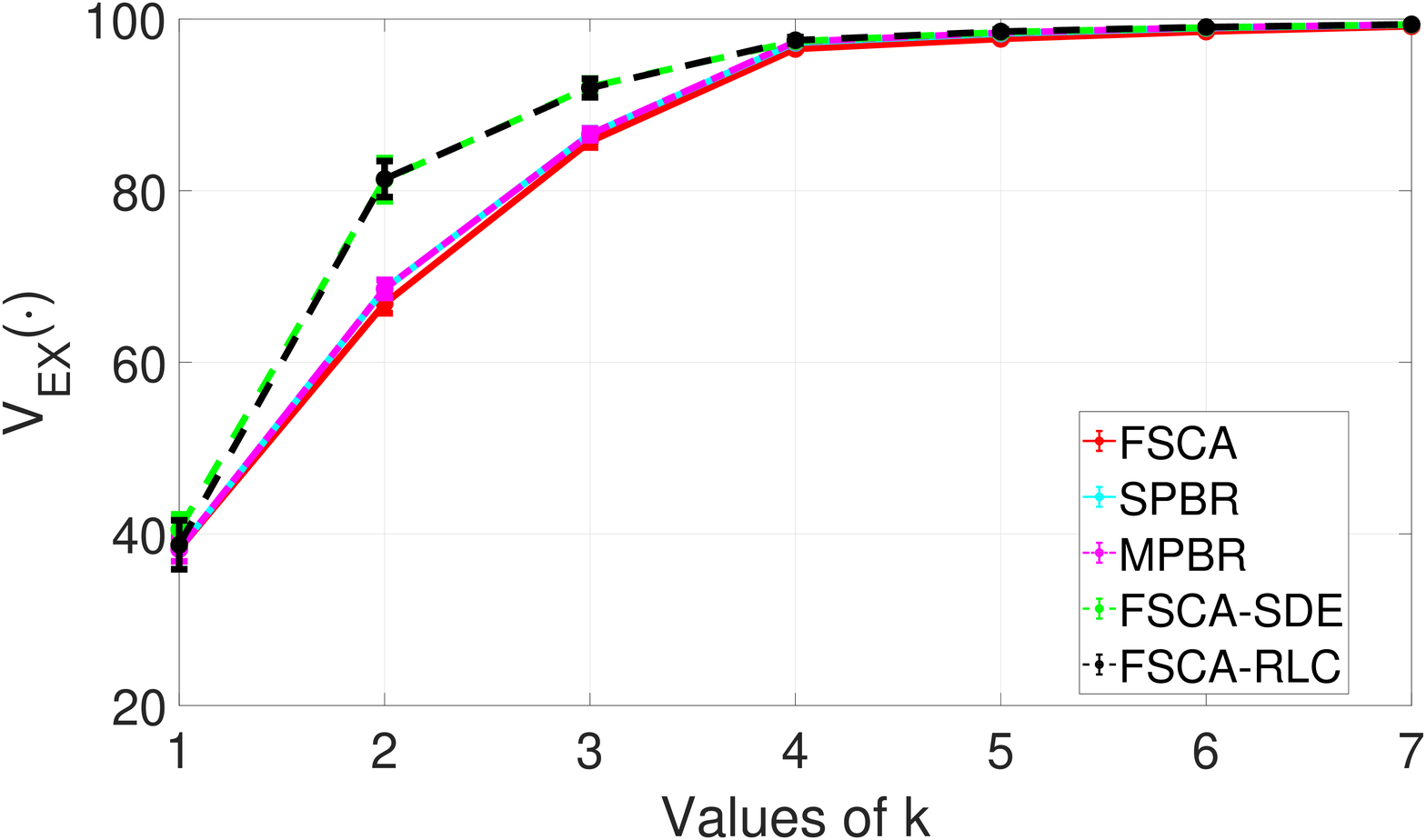} &
\includegraphics[width=0.32\textwidth]{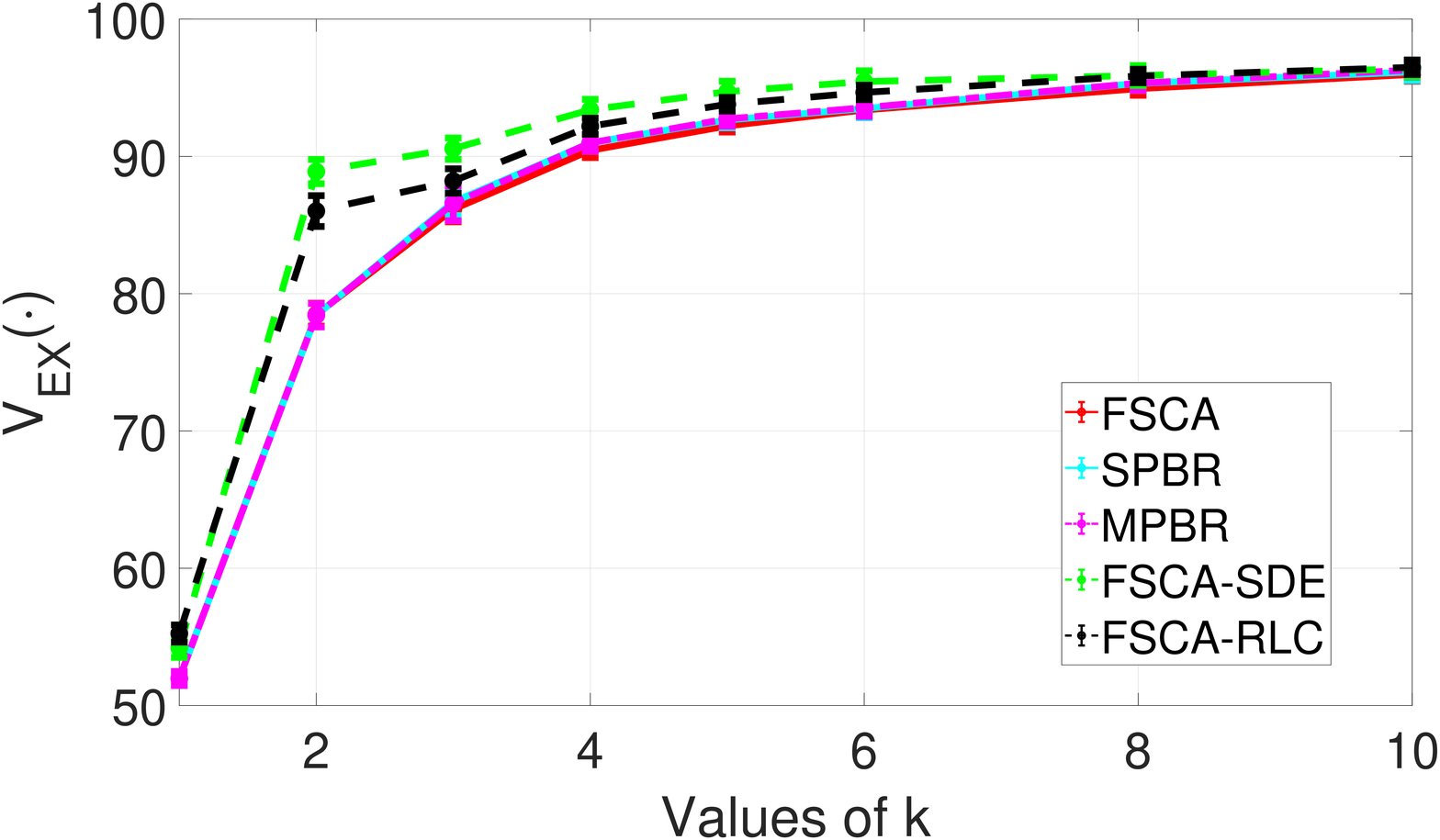} &
\includegraphics[width=0.32\textwidth]{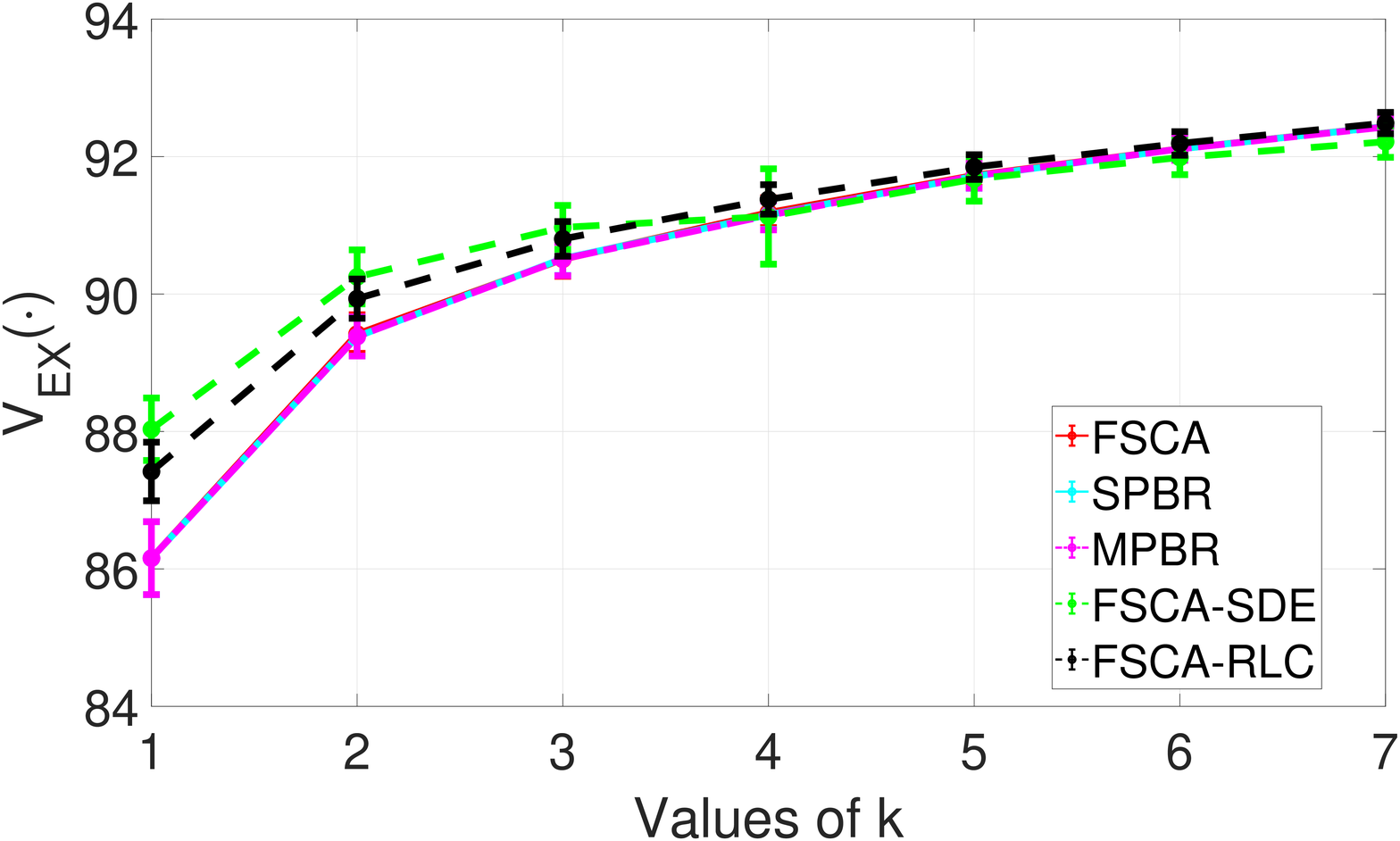}\\ 
(a) Xwafer  & (b) Xchemistry & (c) Xbusiness \\[6pt]
\end{tabular}
\begin{tabular}{ccc}
\includegraphics[width=0.32\textwidth]{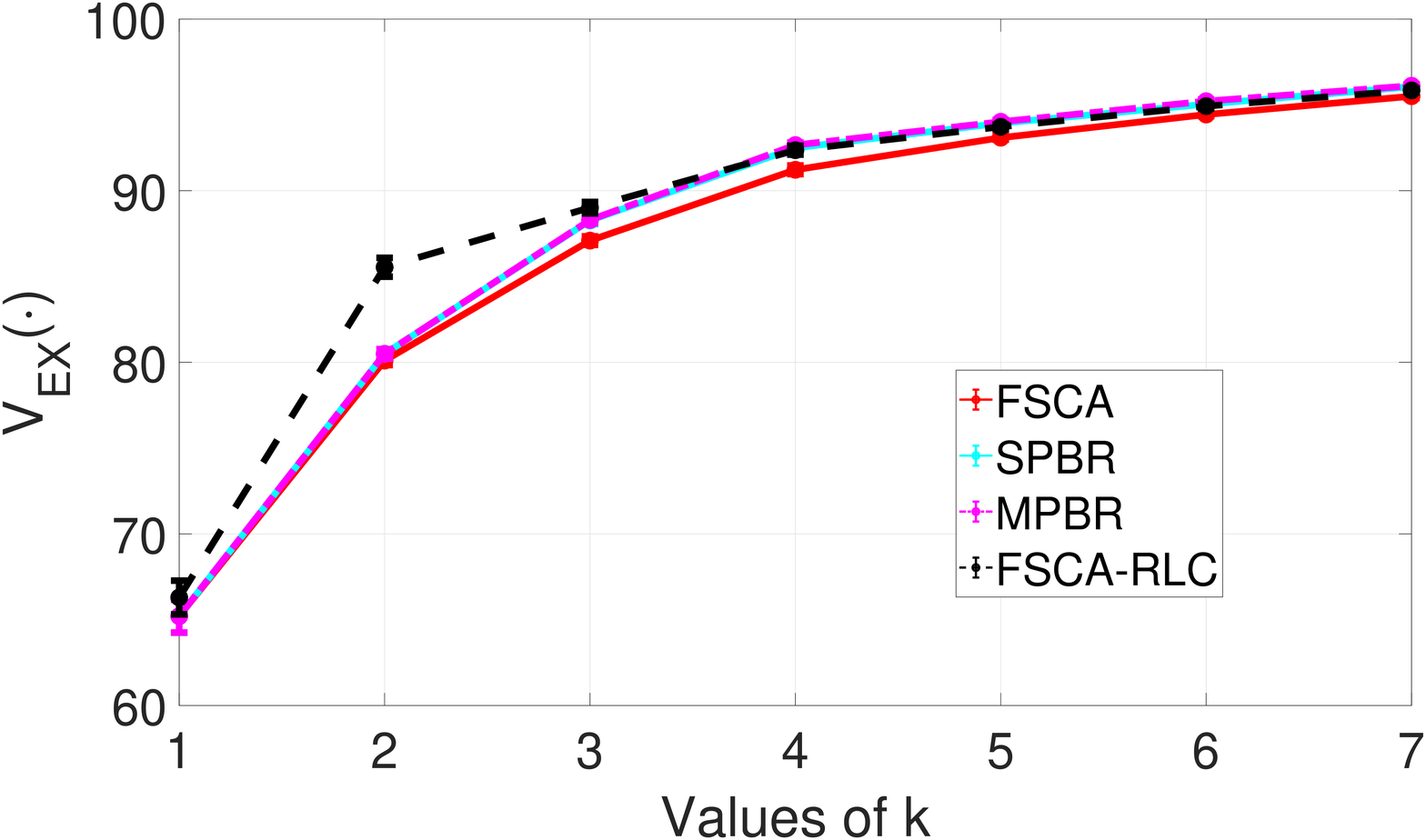} &
\includegraphics[width=0.32\textwidth]{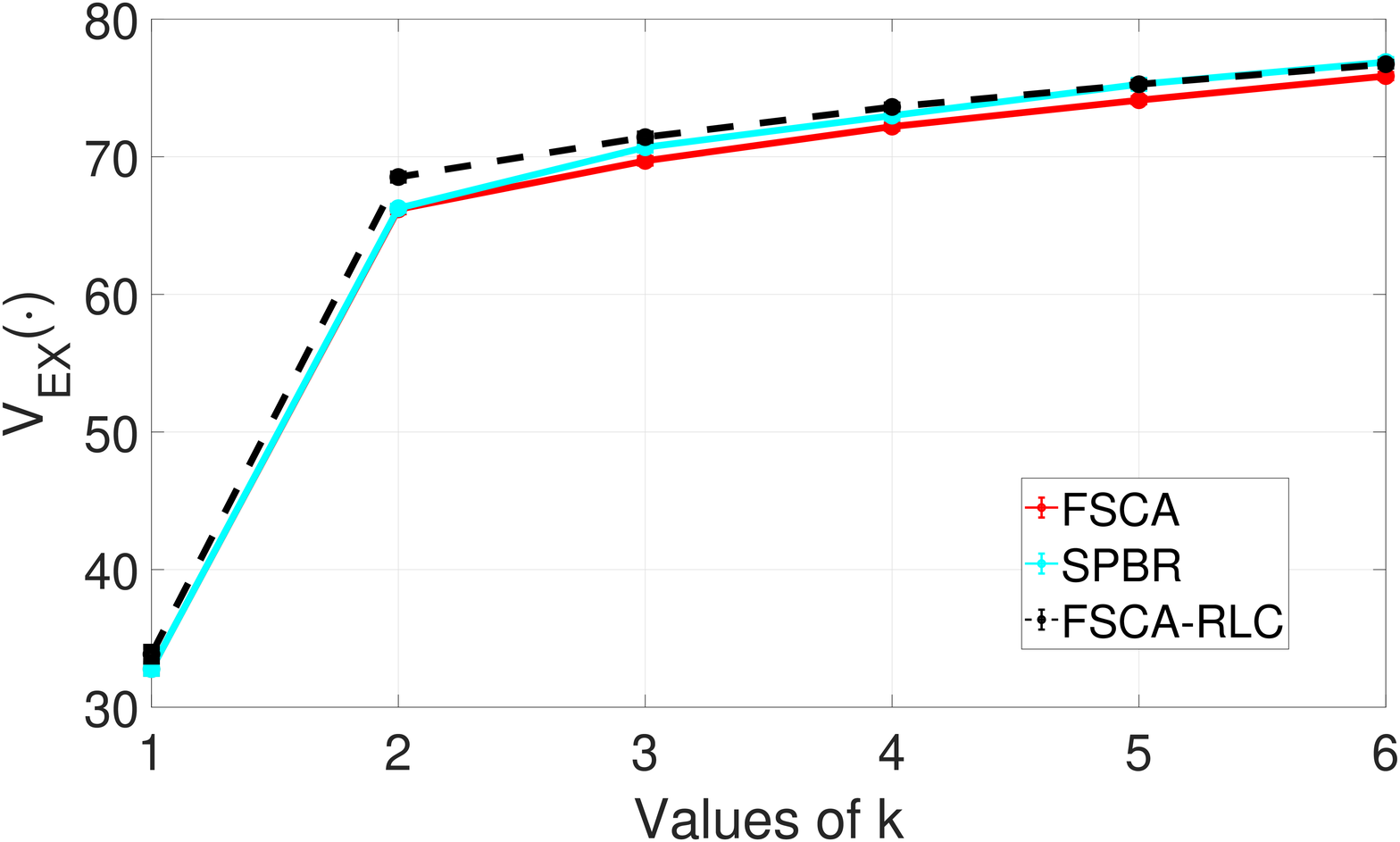} &
\includegraphics[width=0.32\textwidth]{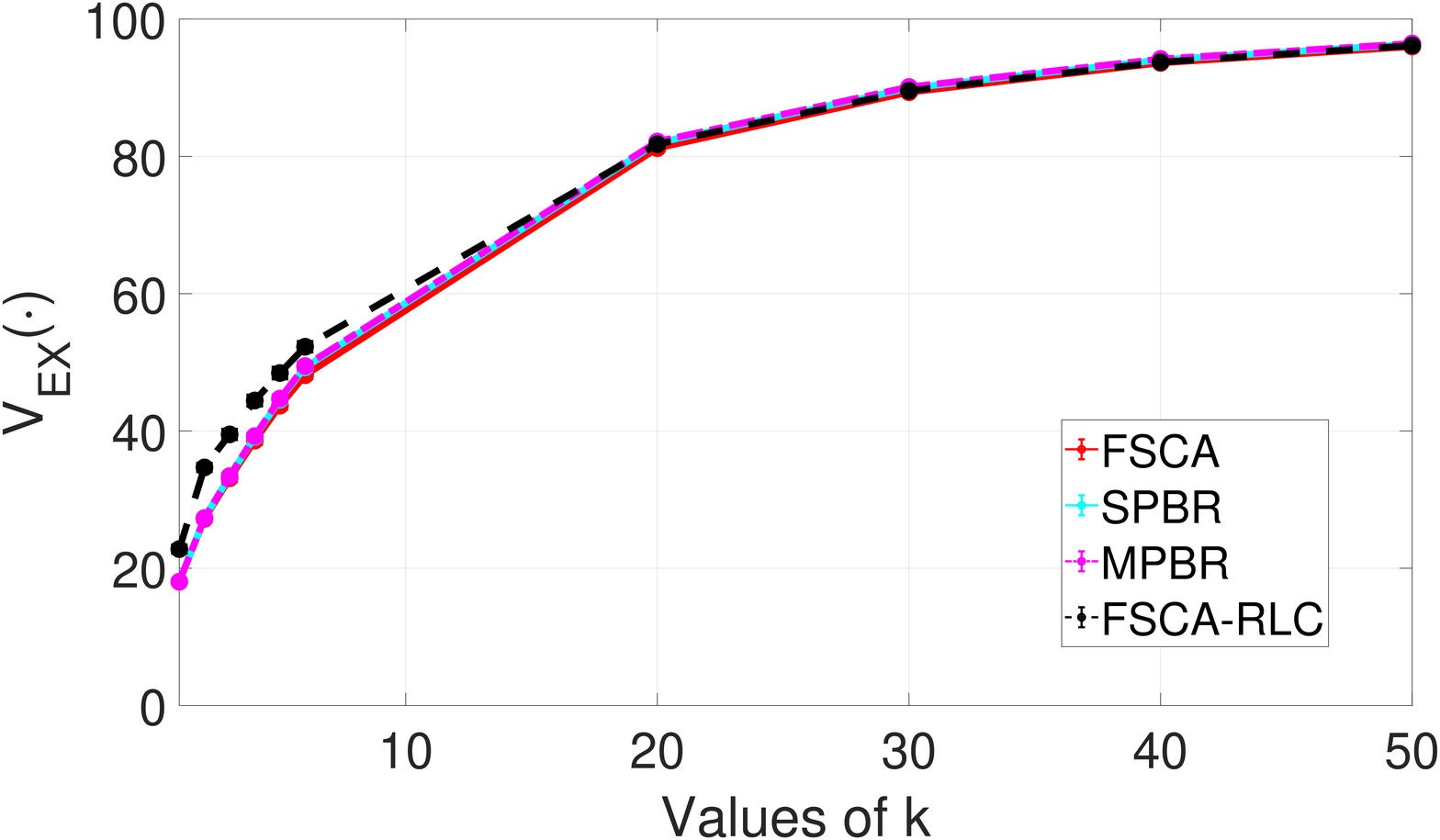} \\
(d) Xsemicon & (e) YaleB & (f) USPS \\[6pt]
\end{tabular}
\caption{Variance explained as a function of $k$. Each line connects the mean values, while the vertical spread is the standard deviation. Table \ref{tab:VarSelSettings} provides details of the simulation setting employed. Missing methods in (c), (d) and (e) have been omitted as they required an excessive training time to be competitive (due to the large value of $v$ and $k_{lin}$).\\}
\label{fig:VarSel-OtherDatasets}
\end{figure*}

To assess the performance of the variable selection algorithms more generally they are further evaluated for six real world datasets, namely, data on weekly purchases of products (Xbusiness \cite{tan2014time}), chemical sensor array data (Xchemistry \cite{vergara2012chemical, rodriguez2014calibration}), the extended Yale face database B (YaleB \cite{GeBeKr01, CHHH07}), the USPS handwritten digit dataset (USPS \cite{cai2010graph, cai2011speed}), and two industrial case studies on semiconductor manufacturing (Xwafer \cite{Prakash2012a} and Xsemicon \cite{puggini2015extreme}). Further details on these datasets and the experimental settings employed are provided in Table \ref{tab:VarSelSettings}. The computation times (relative to FSCA) for each algorithm are reported in Table \ref{tab:VarSel-OtherDatasets-Time} while their performance in terms of variance explained as a function of $k$ is presented in Fig. \ref{fig:VarSel-OtherDatasets}. Whenever results have been omitted for an algorithm it is because it has proven to be too computationally expensive to compute compared to the other algorithms.  

Overall, the most effective variable selection method is FSCA-SDE followed closely by FSCA-RLC, but the former is at least 100 times more computationally complex to train than the latter, hence it was deemed intractable for the largest datasets. Consistent with the finding of \cite{Puggini2017}, MPBR has an accuracy comparable to SPBR and a computation complexity that increases more rapidly with $k$. For the lowest values of $k$ SPBR is the fastest algorithm to compute, but it is less accurate than FSCA-RLC. As $k$ increases, the difference in variance explained between algorithms decreases and for $k \geq 40$ FSCA-RLC becomes faster to compute than SPBR, as observed with YaleB and USPS. This can be explained considering that SPBR has $O(2v^2k^3 + 8vk^3)$ complexity and hence $O(k^3)$ complexity with respect to $k$, whereas the RLC neural network has a complexity of $O(m\theta e)$ where $e$ is the number of epochs required to satisfy early stopping and $\theta$ is the number of weights in the network \cite{Maaten2009}. In these experiments $\theta = kh + h + h\bar{k} + \bar{k} = hk_{lin} + h + k_{lin} - k$, hence the size of the network decreases as $k$ increases, as does the computational complexity per iteration (hypothesizing that $e$ does not increase if the network gets smaller).             

\begin{table*}[!t]
\caption{Variable Selection Simulation Settings Used in Each Case Study. \\ $m_{tr} (\%)$ is the Percentage of the Data Used for Training, `No. MC Sims' is the Number of Monte Carlo \\Simulations Performed, and `Size of $\mathcal{N}$' is the network dimensions in the Form [Input, Hidden Layers, Output].}
\centering
\begin{tabular}{ll|ccc|lcc}
Dataset ($m \times v$) & Source & $m_{tr} (\%)$ & No. MC sims & $\tau$ & Method & Size of $\mathcal{N}$ & Epochs limit \\
\hline
\multirow{2}{*}{Xsynthetic ($500 \times 50$)} & \multirow{2}{*}{Sec. \ref{subsec:var-sel-simulated}} & \multirow{2}{*}{70} & \multirow{2}{*}{100} & \multirow{2}{*}{99} & FSCA-SDE & [$k$, 11 \,21, $v$] & 1000\\
& & & & & FSCA-RLC & [$k$, 6, $\bar{k}]$ & 1000\\
\hline
\multirow{2}{*}{Xwafer ($316 \times 50$)} & \multirow{2}{*}{\cite{Prakash2012a}} & \multirow{2}{*}{70} & \multirow{2}{*}{100} & \multirow{2}{*}{99.9} & FSCA-SDE & [$k$, 15 \,25, $v$] & 1000\\
& & & & & FSCA-RLC & [$k$, 15, $\bar{k}]$ & 1000\\
\hline
\multirow{2}{*}{Xbusiness\footnotemark ($811 \times 52$)} & \multirow{2}{*}{\cite{tan2014time}} & \multirow{2}{*}{70} & \multirow{2}{*}{100} & \multirow{2}{*}{97} & FSCA-SDE & [$k$, 11 \,21, $v$] & 1000\\
& & & & & FSCA-RLC & [$k$, 1, $\bar{k}]$ & 1000\\
\hline
\multirow{2}{*}{Xchemistry\footnotemark ($1586 \times 128$)} & \multirow{2}{*}{\cite{vergara2012chemical, rodriguez2014calibration}} & \multirow{2}{*}{40} & \multirow{2}{*}{30} & \multirow{2}{*}{99} & FSCA-SDE & [$k$, 13 \,18, $v$] & 20\\
& & & & & FSCA-RLC & [$k$, 3, $\bar{k}]$ & 1000\\
\hline
\multirow{2}{*}{Xsemicon ($2194 \times 2046$)} & \multirow{2}{*}{\cite{puggini2015extreme}} & \multirow{2}{*}{50} & \multirow{2}{*}{50} & \multirow{2}{*}{97} & FSCA-SDE & - & -\\
& & & & & FSCA-RLC & [$k$, 3, $\bar{k}]$ & 1000\\
\hline
\multirow{2}{*}{YaleB\footnotemark ($2414 \times 1024$)} & \multirow{2}{*}{\cite{GeBeKr01, CHHH07}} & \multirow{2}{*}{45} & \multirow{2}{*}{60} & \multirow{2}{*}{97} & FSCA-SDE & - & -\\
& & & & & FSCA-RLC & [$k$, 5, $\bar{k}]$ & 8\\
\hline
\multirow{2}{*}{USPS\footnotemark[\value{footnote}] ($9298 \times 256$)} & \multirow{2}{*}{\cite{cai2010graph, cai2011speed}} & \multirow{2}{*}{15} & \multirow{2}{*}{70} & \multirow{2}{*}{97} & FSCA-SDE & - & -\\
& & & & & FSCA-RLC & [$k$, 5, $\bar{k}]$ & 20\\
\hline
\end{tabular}
\label{tab:VarSelSettings}
\end{table*}
\footnotetext[2]{\tiny Downloaded from: \emph{https://archive.ics.uci.edu/ml/datasets/Sales\_Transactions\_Dataset\_Weekly\#} (when you paste the link in the search engine, manually add the 3 underscores).}
\footnotetext[3]{\tiny Downloaded from: \emph{https://archive.ics.uci.edu/ml/datasets/Gas+Sensor+Array+Drift+Dataset+at+Different\\+Concentrations} (only the batch number 3 was considered).}
\footnotetext[4]{\tiny Downloaded from: \emph{http://www.cad.zju.edu.cn/home/dengcai/Data/data.html}}

\begin{table*}[!t]
\centering
\caption{Mean(Standard Deviation) of the Computation Time for Each Method (Expressed as a\\ Fraction of the  Computation Time for FSCA) for Different Values of $k$ for Each Dataset. \\The Best Performing Methods for Each Dataset and Each Value of $k$ are Highlighted in Bold.\\}
\begin{tabular}{c|cc|cc|cc|cc|cc|cc}

\multirow{2}{*}{{Method}}& \multicolumn{2}{|c|}{{Xsynthetic}} &\multicolumn{2}{|c|}{{Xchemistry}} & \multicolumn{2}{|c|}{{Xbusiness}} &\multicolumn{2}{|c|}{{Xsemicon}} & \multicolumn{2}{|c|}{{YaleB}} & \multicolumn{2}{|c}{{USPS}}\\
& $k$ & Mean(std) & $k$ & Mean(std) & $k$ & Mean(std) & $k$ & Mean(std) & $k$ & Mean(std) & $k$ & Mean(std) \\
\hline
\multirow{3}{*}{SPBR} & 2 & $\bm{4.4(3.8)}$ & 2 & $\bm{5.8 (2.9)}$ & 2 & $\bm{4.5 (2.5)}$ & 2 & $\bm{2.5 (2.3)}$ & 2 & $\bm{10.0 (2.4)}$ & 2 & $\bm{6.0 (3.0)}$\\
                                    & 5 & $\bm{7.2(5.3)}$ & 8 & $\bm{16.5 (2.0)}$ & 5 & $\bm{8.1 (8.5)}$ & 5 & 4.4(2.2) & 40 & 37.1(20.0) & 30 & 34.1(2.3) \\
                                    & 8 & $\bm{9.7(11.7)}$ & 16 & 33.2(8.2) & 7 & $\bm{9.7(13.1)}$ & 7 & 5.1(1.8) & 80 & 54.7(30.9) & 50 & 44.1(3.9)\\
\hline
\multirow{3}{*}{MPBR} & 2 & 6.2(2.4) & 2 & 7.4(12.0) & 2 & 4.9 (3.7) & 2 & 4.0(3.0) & - & - & 2 & 9.6(12.1)\\
                                     & 5 & 13.2(10.1) & 8 & 48.6(75.9) & 5 & 13.4(47.3) & 5 & 12.6(213.9) & - & - & 30 & 131.4(168.6)\\
                                     & 8 & 27.0(77.3) & 16 & 100.8(130.5) & 7 & 17.8(112.8) & 7 & 16.8(120.4) & - & - & 50 & 194.6(247.1)\\
\hline
\multirow{3}{*}{\pbox{20cm}{FSCA-SDE \\ $\div10^{4}$}} & 2 & 3.6(0.6) & 2 & 15.4(3.6) & 2 & 3.6(2.7) & - & - & - & - & - & -\\
                                          & 5 & 1.9(2.7) & 8 & 4.4(0.9) & 5 & 1.4(3.2) & - & - & - & - & - & -\\
                                          & 8 & 1.5(3.6) & 16 & 2.2(0.5) & 7 & 1.0(3.3) & - & - & - & - & - & -\\
\hline
\multirow{3}{*}{FSCA-RLC} & 2 & 216.2(43.4) & 2 & 211.5(857.3) & 2 & 193.0(29.0) & 2 & 5.0(7.2) & 2 & 64.9(36.0) & 2 & 224.0(48.4)\\
                                   & 5 & 109.9(332.3) & 8 & 40.8(576.2) & 5 & 81.1(64.6) & 5 & $\bm{2.1 (6.4)}$ & 40 & $\bm{2.9 (5.1)}$ & 30 & $\bm{8.1 (2.6)}$\\
                                   & 8 & 49.6(46.7) & 16 & $\bm{9.0 (12.3)}$ & 7 & 59.7(135.6) & 7 & $\bm{1.4 (2.4)}$ & 80 & $\bm{1.3 (3.4)}$ & 50 & $\bm{1.9 (1.2)}$\\
\hline
\end{tabular}
\label{tab:VarSel-OtherDatasets-Time}
\end{table*}

\textbf{Extended analysis of Xwafer:} 
The development of RLC was initially motivated by a semiconductor manufacturing spatial sampling design problem, as exemplified by the Xwafer dataset \cite{Prakash2012a, mcloone2018optimising}. Here, each column of the data matrix represents the measurements of a process parameter, such as etch depth or deposition layer thickness,  at a specific wafer site (location on the wafer surface) for a batch of wafers. The objective is to identify the minimum number of measurement sites needed to accurately monitor the full wafer surface, or equivalently, to find the minimum subset of columns of the matrix that can achieve a desired level of accuracy in terms of reproducing the full matrix.  An important requirement for practical application is to have a relatively low computation time for the algorithm performing site selection and surface reconstruction. Therefore, complex deep learning techniques are not suitable in this context. 

\begin{figure}
\centering
\includegraphics[width=0.5\textwidth]{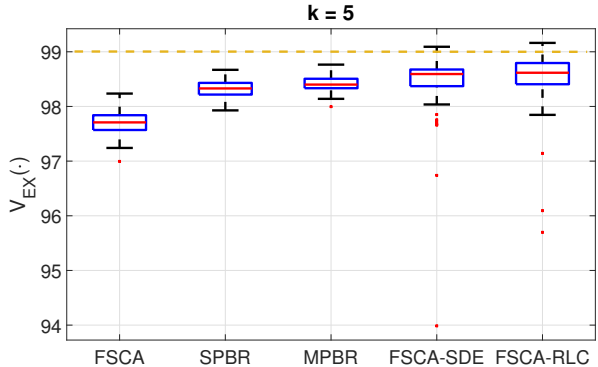}
\caption{Boxplot of the distribution of $V_{EX}$ achieved with each method for $k$ = 5 on Xwafer. The target accuracy is indicated by the dashed line.}
\label{fig:X50sitesBoxplot}
\end{figure}

The site selection problem has already been addressed in \cite{Prakash2012a, mcloone2018optimising} and the state-of-the-art is defined by the SPBR and MPBR proposed in \cite{Puggini2017}, where for a target surface reconstruction accuracy of $99\%$ or greater they yield a minimum measurement plan of 6 sites in contrast to 7 sites with FSCA. To assess the performance of FSCA-RLC and FSCA-SAE for this problem Table \ref{tab:VarSelWafer} focuses on the results for $k = 5$ and $k = 6$ since these are the minimum values of $k$ required to achieve the desired reconstruction accuracy with at least one of the algorithms. FSCA-RLC yields the best performance on average and exceeds the 99\% threshold more frequently than the other methods, with FSCA-SDE a close second. SPBR is the least computationally complex of the 4 algorithms and, along with MPBR, exhibits the smallest standard deviation in performance over the 100 Monte Carlo simulations. The situation for $k = 5$ is further clarified with the boxplot in Fig. \ref{fig:X50sitesBoxplot} where the 99\% threshold is indicated by the yellow dashed line. Note that, even though FSCA-RLC does not employ greedy layer-wise pre-training as part of the learning process, it achieves comparable robustness to overfitting to FSCA-SDE, with the added benefit of being more than 150 times faster to train.  



\begin{table}
\caption{Mean(Standard Deviation) of $V_{EX}$ and Normalized Training Time for Xwafer When $k$=5 and $k$=6. Parameter $p_{99\%}$ is the Probability of Exceeding the $99\%$ $V_{EX}$ Threshold.}
\centering
\begin{tabular}{llcc}
Technique & $V_{EX}$: Mean(std)   & $p_{99\%}$ & Training time: Mean(std) \\
\hline
SPBR($k$=5) & 98.32(0.15) & 0 & $\bm{17.9(2.7)}$\\
MPBR($k$=5) & 98.40(0.13) & 0 & 51.7(61.3)\\
FSCA-SDE($k$=5) & 98.46(0.57) & 0.02 & 16,168(65,643)\\
FSCA-RLC($k$=5) & $\bm{98.55(0.48)}$ & $\bm{0.04}$ & 104.6(242.3)\\
\hline
SPBR($k$=6) & 98.96(0.11) & 0.37 & $\bm{19.0(1.9)}$\\
MPBR($k$=6) & 99.02(0.07) & 0.59 & 53.4(106.1)\\
FSCA-SDE($k$=6) & 99.02(0.33) & 0.64 & 14,778(48,594)\\
FSCA-RLC($k$=6) & $\bm{99.07(0.20)}$ & $\bm{0.68}$ & 93.8(251.6)\\
\hline
\end{tabular}
\label{tab:VarSelWafer}
\end{table} 


\section{Results for Unsupervised Dimensionality Reduction}
\label{sec:DimRed}
In this section three unsupervised dimensionality reduction methods are compared: PCA, PCA-SAE and PCA-RLC. The experimental setup and software tools employed are as introduced in Section \ref{sec:VarSel}. The MATLAB code developed for the experiments is available on GitHub$^1$. Four datesets are used as case studies, namely, Xsyntethic, Xwafer, Xchemistry and USPS. For each one three different PCA-SAE and PCA-RLC training scenarios are investigated:
\vspace{1.0 mm}
\newline
\textbf{\emph{Scenario A}}: In addition to employing the settings detailed in Table \ref{tab:DimRedSettings}, the algorithms are implemented in accordance with Algorithms \ref{Alg:RLC} and \ref{Alg:PCAsae} with different randomly generated training and test datasets used for each Monte Carlo simulation; 
\vspace{1.0 mm}
\newline
\textbf{\emph{Scenario B}}: Here the same training and test datasets are used in all simulations. Hence, any variation in $V_{EX}$ will be a consequence of the neural network training process only. In addition, greedy layer-wise pre-training is not performed with PCA-SAE. 
\vspace{1.0 mm}
\newline
\textbf{\emph{Scenario C}}: This is an extension of \emph{Scenario B} where a network size equivalence condition (\ref{Eq:equivalenceCond}) is imposed on PCA-SAE and PCA-RLC, as described later.
\vspace{1.0 mm}
\begin{table*}[!t]
\caption{Simulation Settings Used in the Dimensionality Reduction Experiments. $m_{tr} (\%)$ is the Percentage of Data Used for Training,\\  `No. MC Sims' is the Number of Monte Carlo Simulations Performed, and `Size of $\mathcal{N}$' is the Network Dimensions in the Form [Input, Hidden Layers, Output] \emph{Without} the Equivalence Condition Applied  $||$ \emph{With} the Equivalence Condition Applied.}
\centering
\begin{tabular}{l|ccc|ccc}
Dataset ($m \times v$) & $m_{tr} (\%)$ & No. MC sims & $\tau$ & Method & Size of $\mathcal{N}$ & Epochs limit \\
\hline
\multirow{2}{*}{Xsynthetic ($500 \times 50$)} & \multirow{2}{*}{70} & \multirow{2}{*}{70} & \multirow{2}{*}{99} & PCA-SAE & [$k_{lin}$, 24 \, $k$ \, 24, $k_{lin}$] $||$ [$k_{lin}$, $k$, $k_{lin}$] & 1000\\
& & & & PCA-RLC & [$k$, 4, $\bar{k}$] $||$ [$k$, 2$k$, $\bar{k}$] & 1000\\
\hline
\multirow{2}{*}{Xwafer ($316 \times 50$)} & \multirow{2}{*}{70} & \multirow{2}{*}{70} & \multirow{2}{*}{99} & PCA-SAE & [$k_{lin}$, 30 \, $k$ \, 30, $k_{lin}$] $||$ [$k_{lin}$, $k$, $k_{lin}$] & 1000\\
& & & & PCA-RLC & [$k$, 10, $\bar{k}$] $||$ [$k$, 2$k$, $\bar{k}$] & 1000\\
\hline
\multirow{2}{*}{Xchemistry ($1586 \times 128$)} & \multirow{2}{*}{50} & \multirow{2}{*}{50} & \multirow{2}{*}{98} & PCA-SAE & [$k_{lin}$, 24 \, $k$ \, 24, $k_{lin}$] $||$ [$k_{lin}$, $k$, $k_{lin}$] & 1000\\
& & & & PCA-RLC & [$k$, 4, $\bar{k}]$ $||$ [$k$, 2$k$, $\bar{k}$] & 1000\\
\hline
\multirow{2}{*}{USPS ($9298 \times 256$)} & \multirow{2}{*}{15} & \multirow{2}{*}{30} & \multirow{2}{*}{97} & PCA-SAE & [$k_{lin}$, 33 \, $k$ \, 33, $k_{lin}$] $||$ [$k_{lin}$, $k$, $k_{lin}$] & 25\\
& & & & PCA-RLC & [$k$, 3, $\bar{k}$] $||$ [$k$, 2$k$, $\bar{k}$] & 1000\\
\hline
\end{tabular}
\label{tab:DimRedSettings}
\end{table*} 

\begin{figure*}
\centering
\begin{tabular}{cccc}
\includegraphics[width=0.23\textwidth]{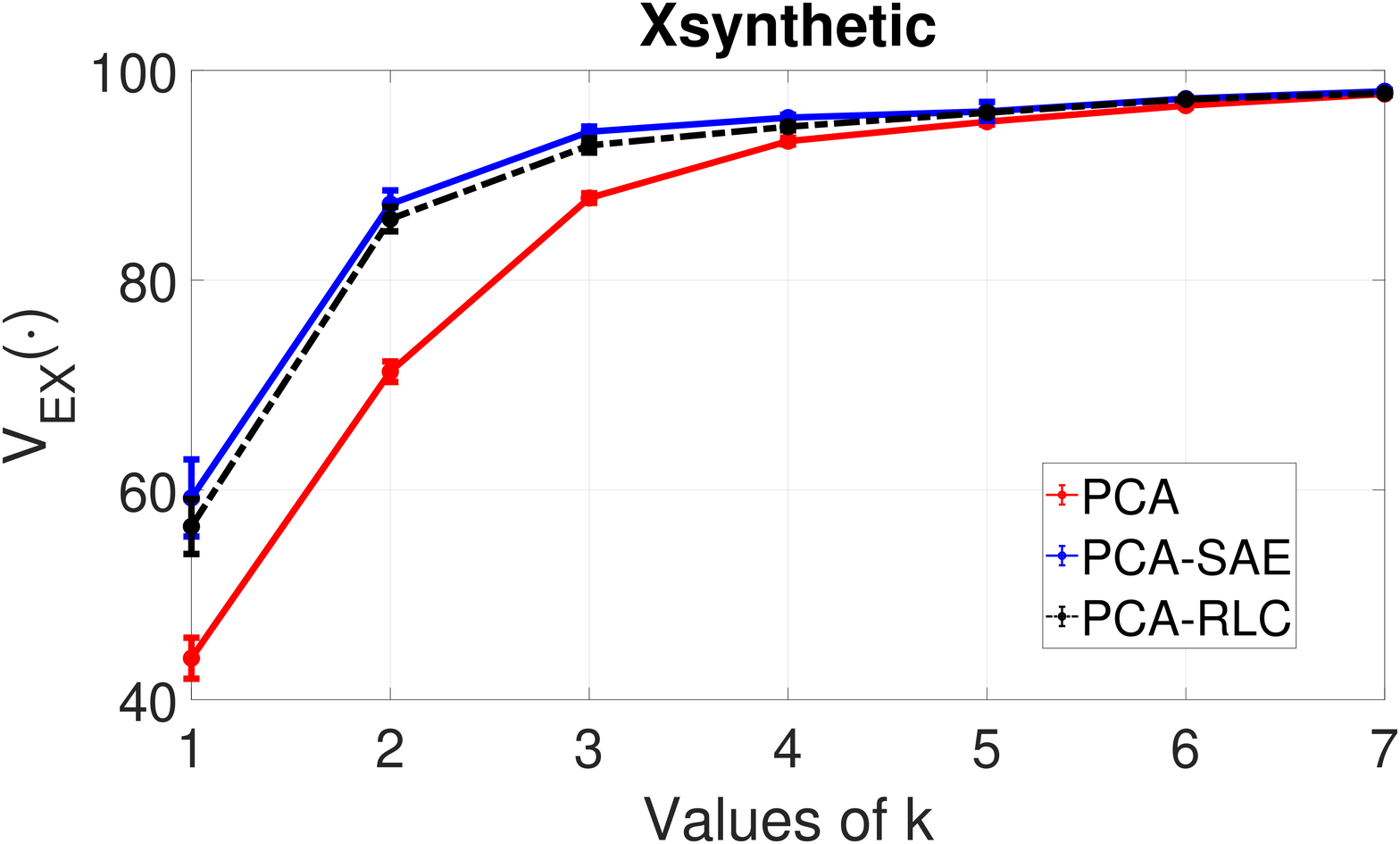} &
\includegraphics[width=0.23\textwidth]{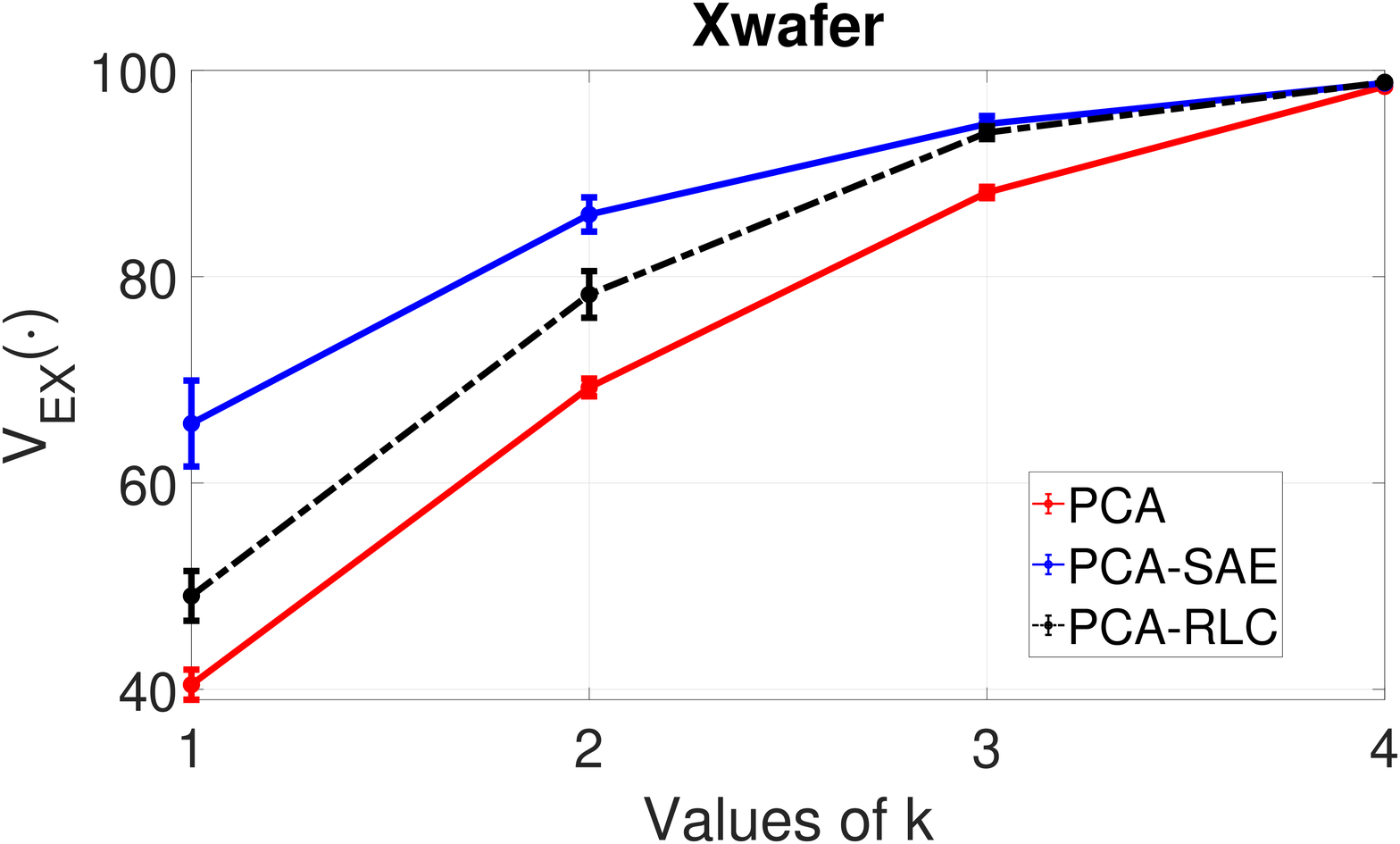} &
\includegraphics[width=0.23\textwidth]{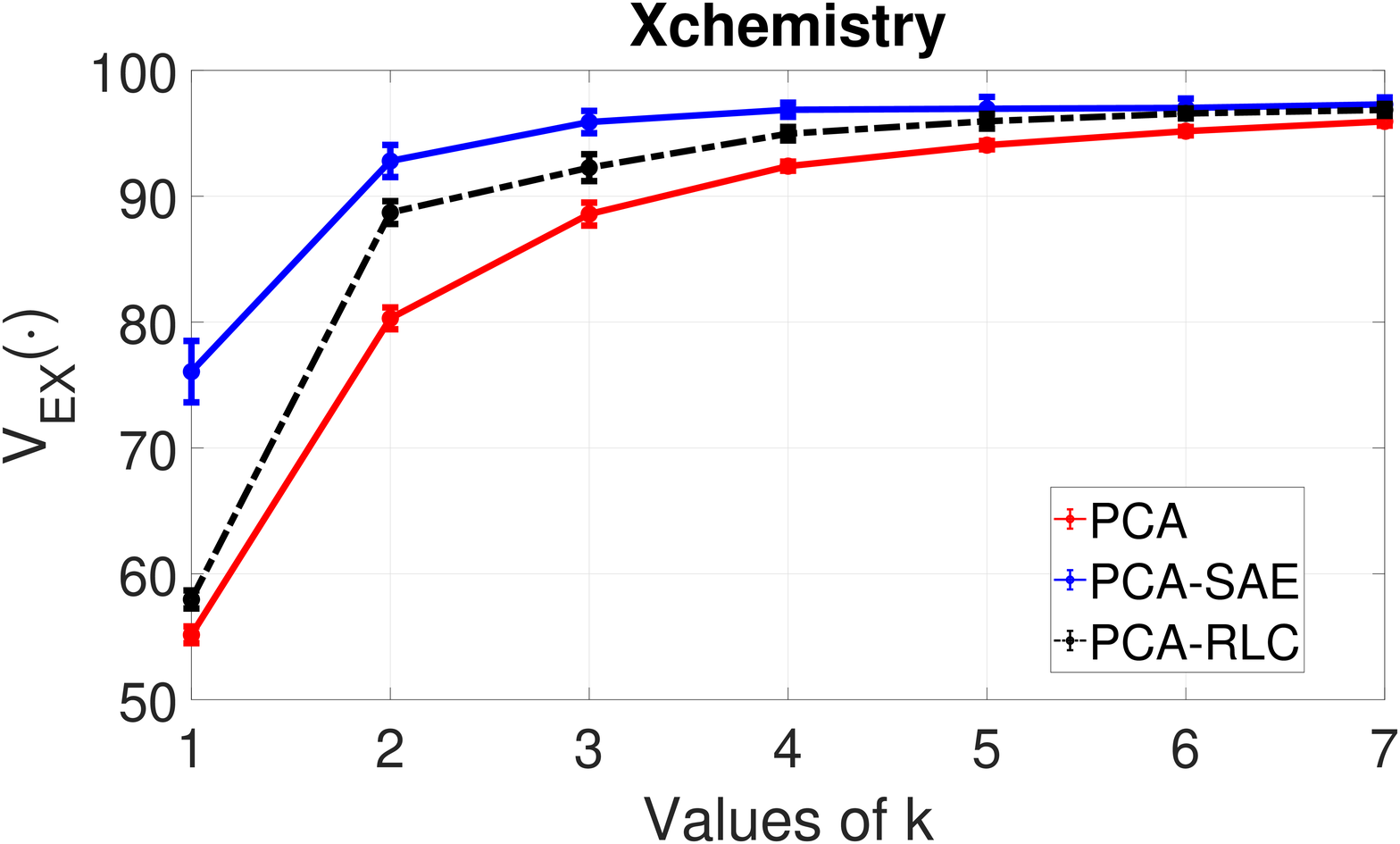} &
\includegraphics[width=0.23\textwidth]{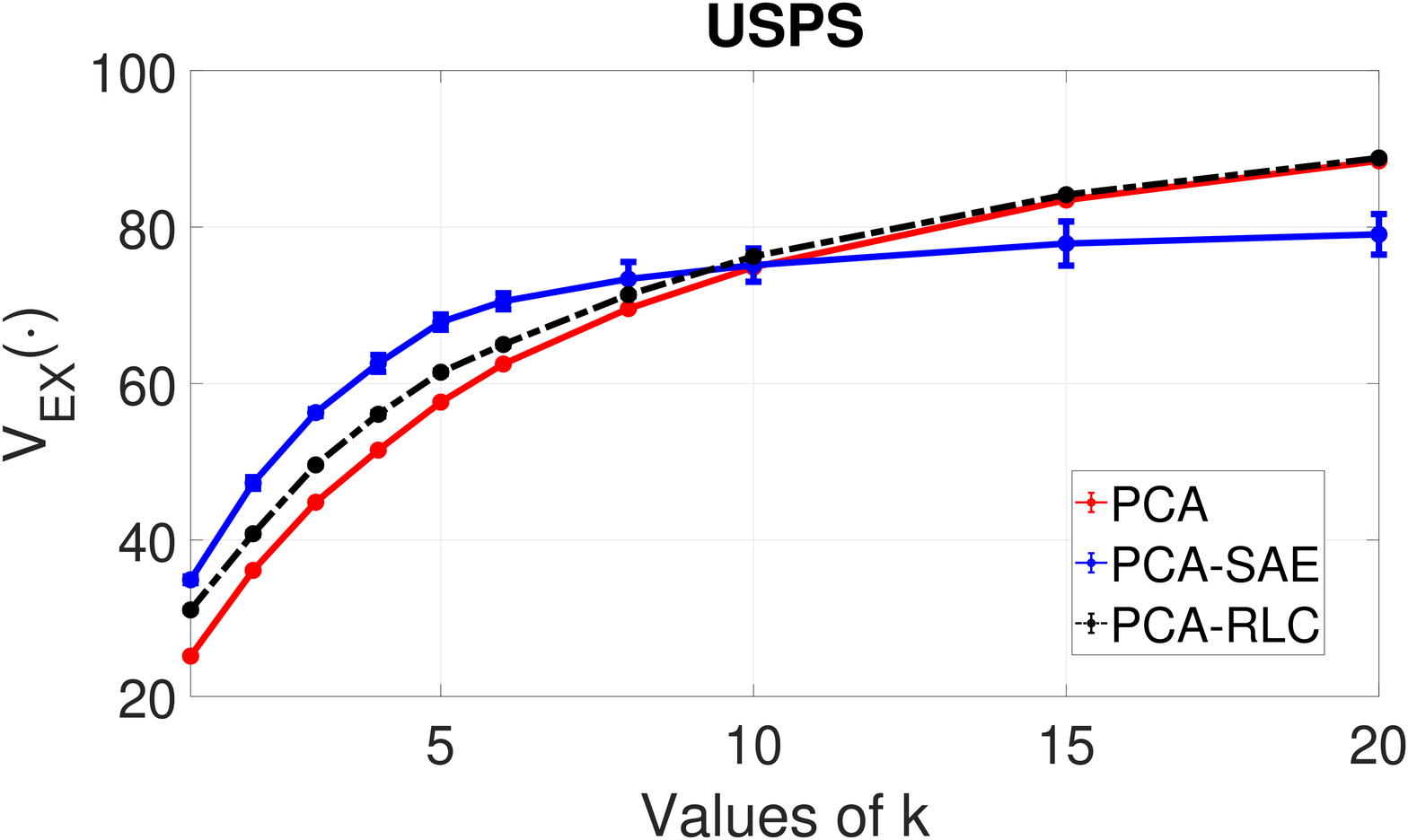} \\
\end{tabular}
\begin{tabular}{c}
\small (a) [\emph{Scenario A}] Training/test sets randomly generated for each Monte Carlo simulation.\\[6pt]
\end{tabular}
\begin{tabular}{cccc}
\includegraphics[width=0.23\textwidth]{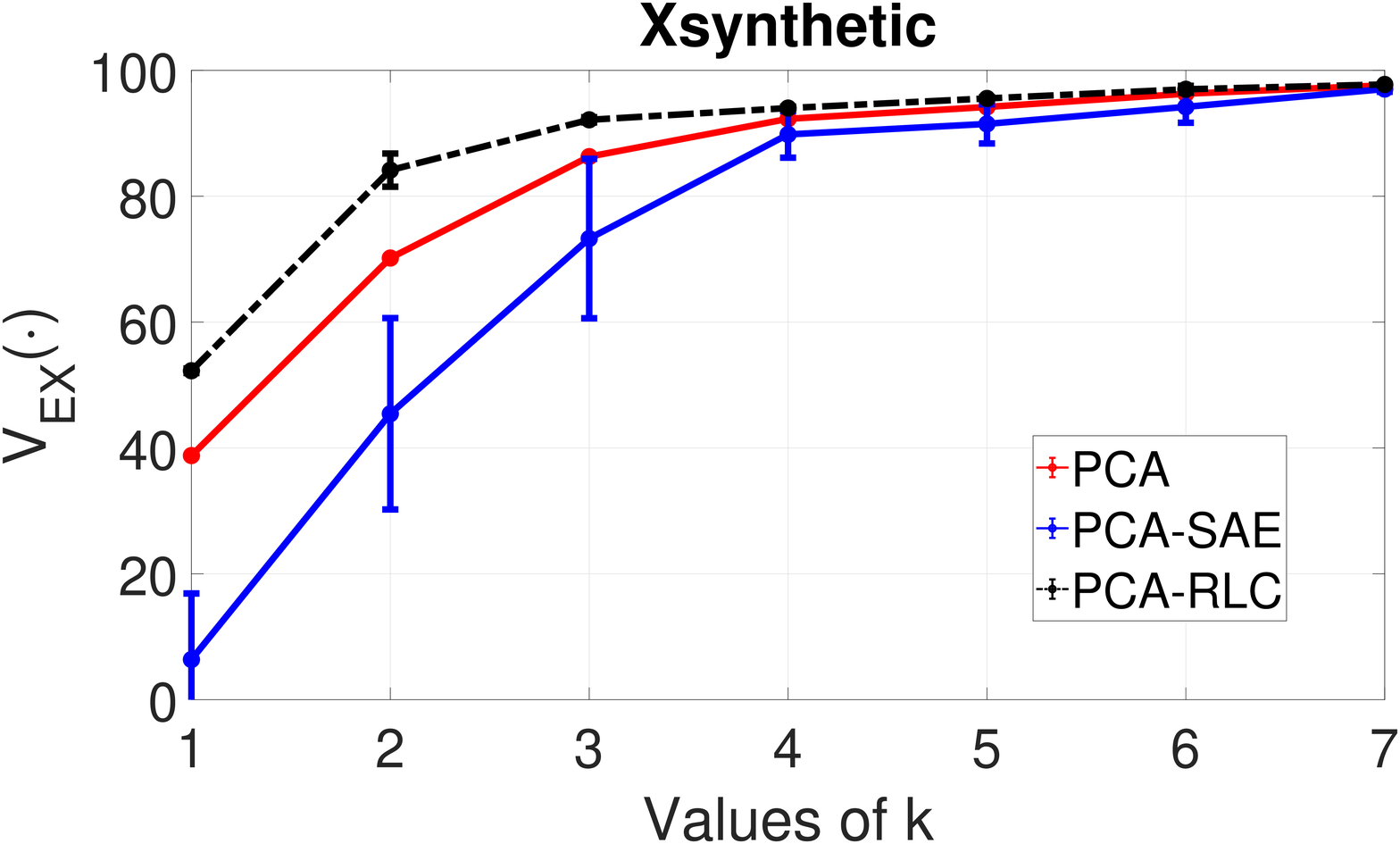} &
\includegraphics[width=0.23\textwidth]{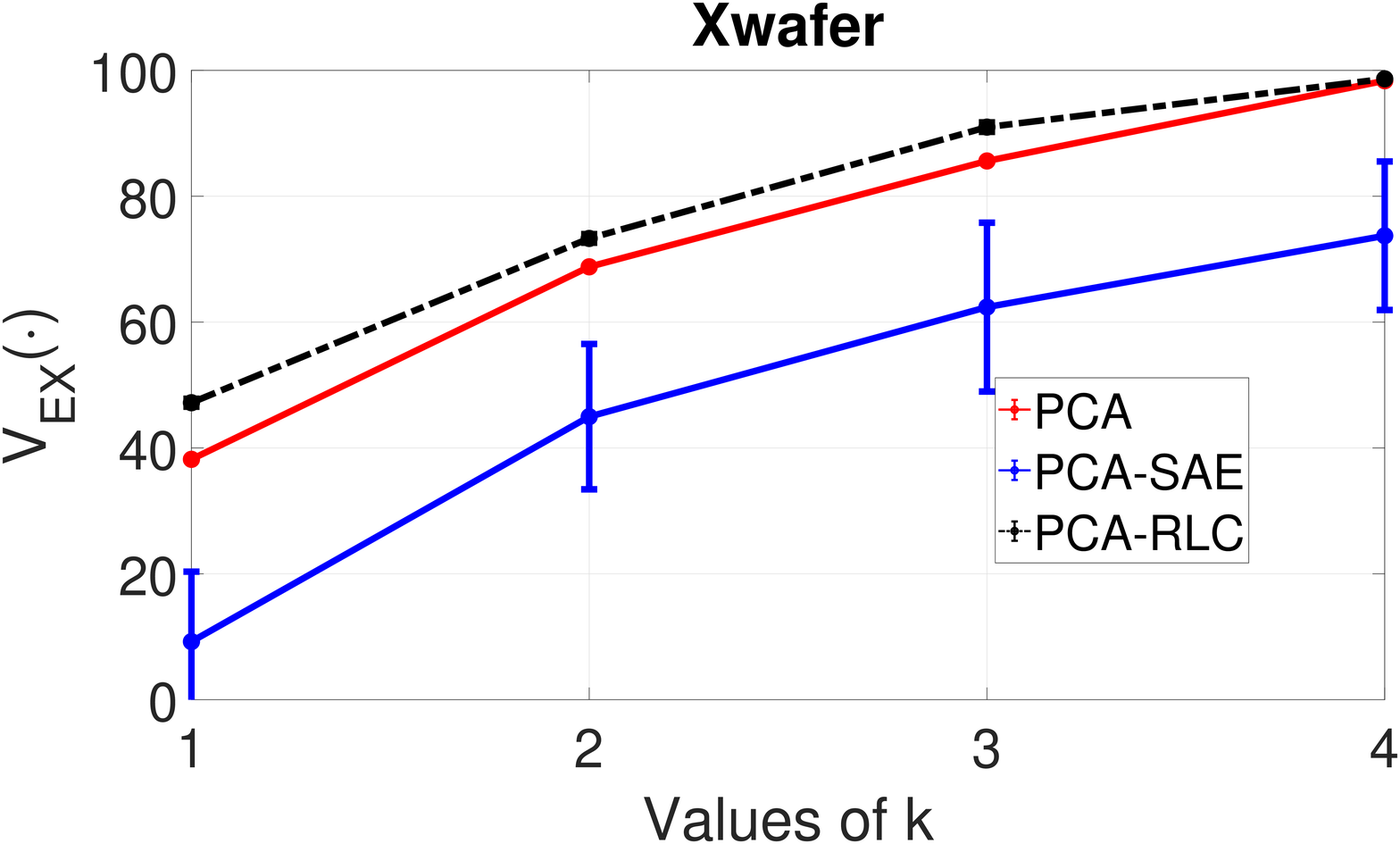} &
\includegraphics[width=0.23\textwidth]{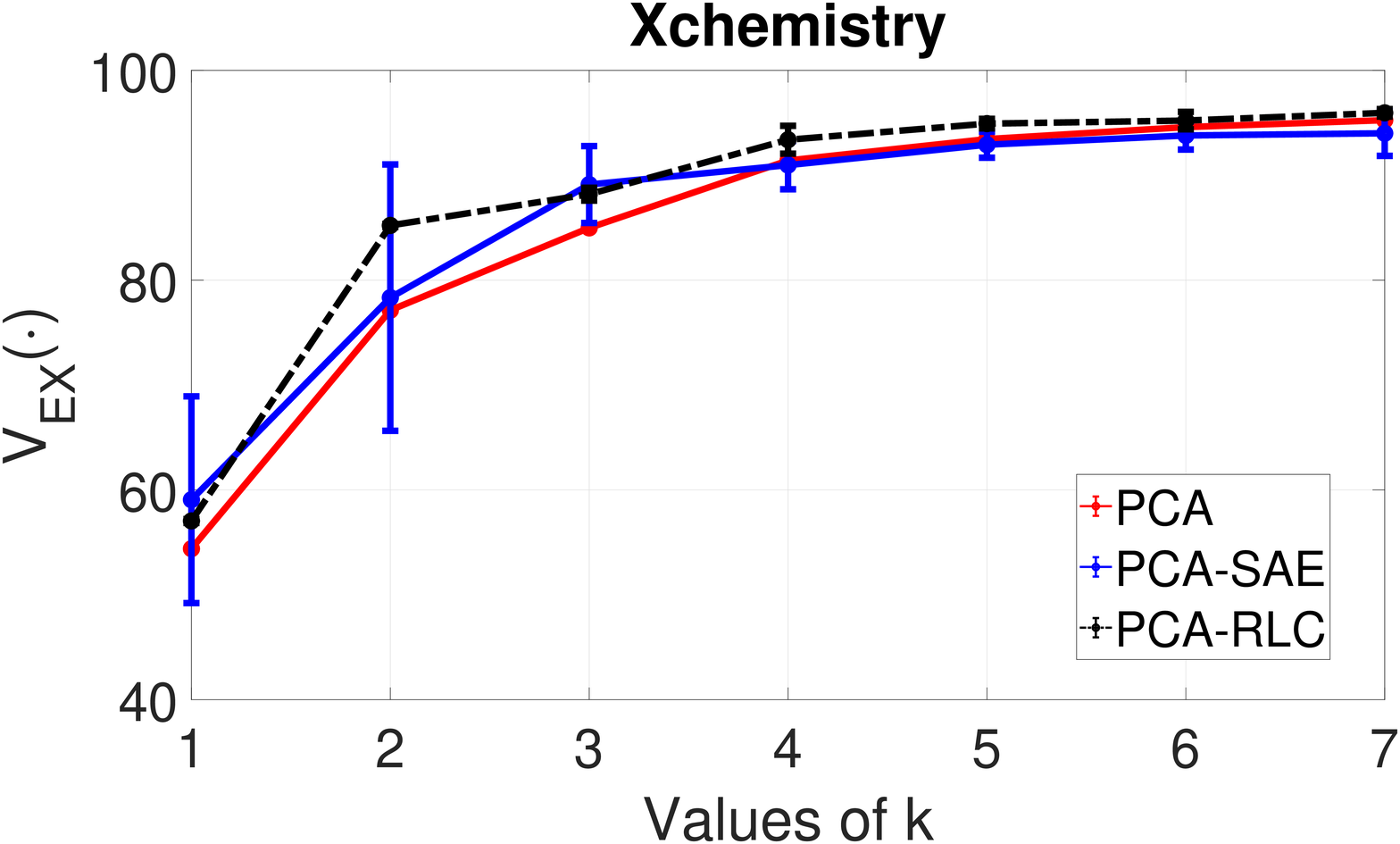} &
\includegraphics[width=0.23\textwidth]{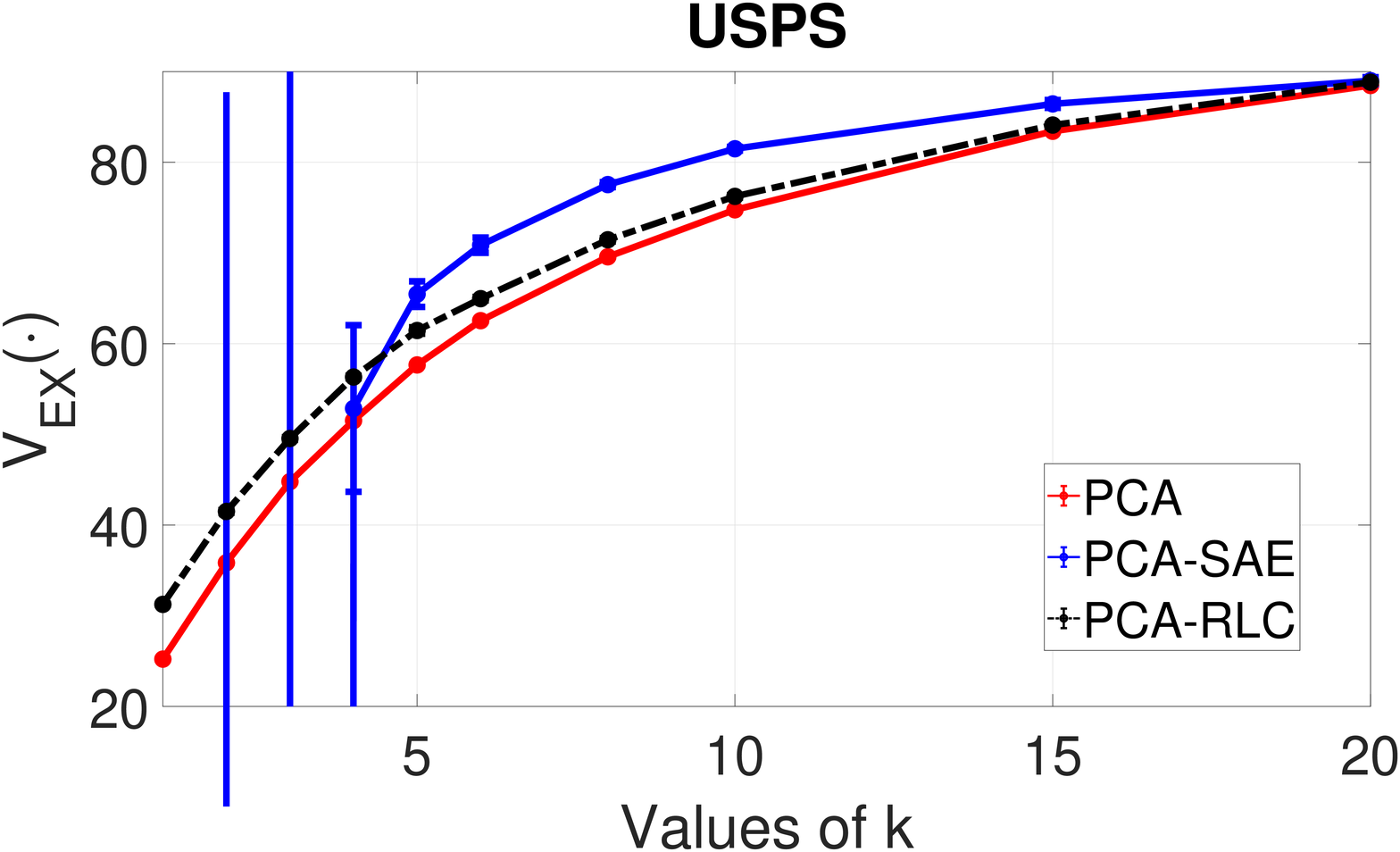} \\ 
\end{tabular}
\begin{tabular}{c}
\small (b) [\emph{Scenario B}] The same training/test sets used in all Monte Carlo simulations and PCA-SAE\\ \small trained \emph{without} using greedy layer-wise pre-training.\\[6pt]
\end{tabular}
\begin{tabular}{cccc}
\includegraphics[width=0.23\textwidth]{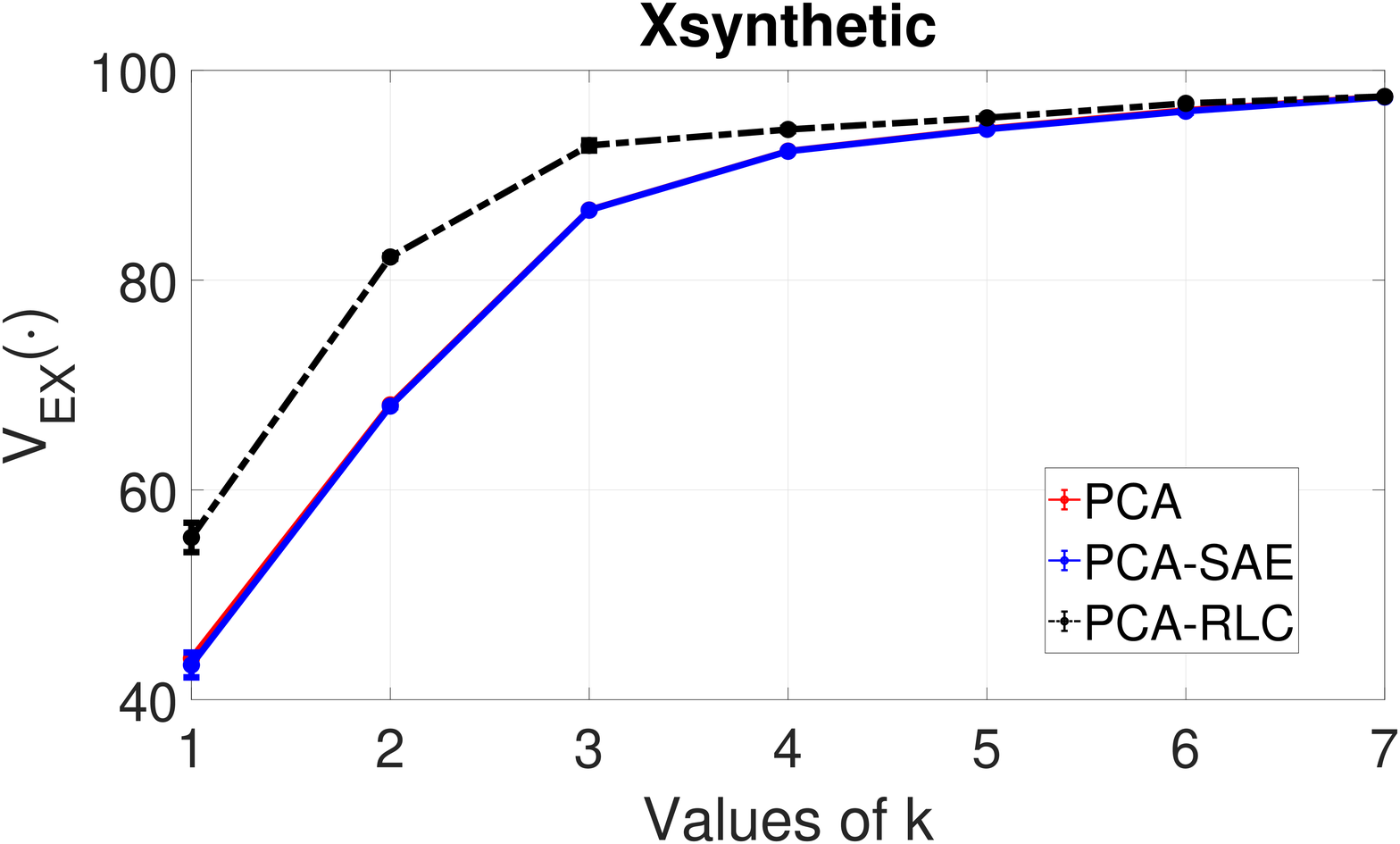} &
\includegraphics[width=0.23\textwidth]{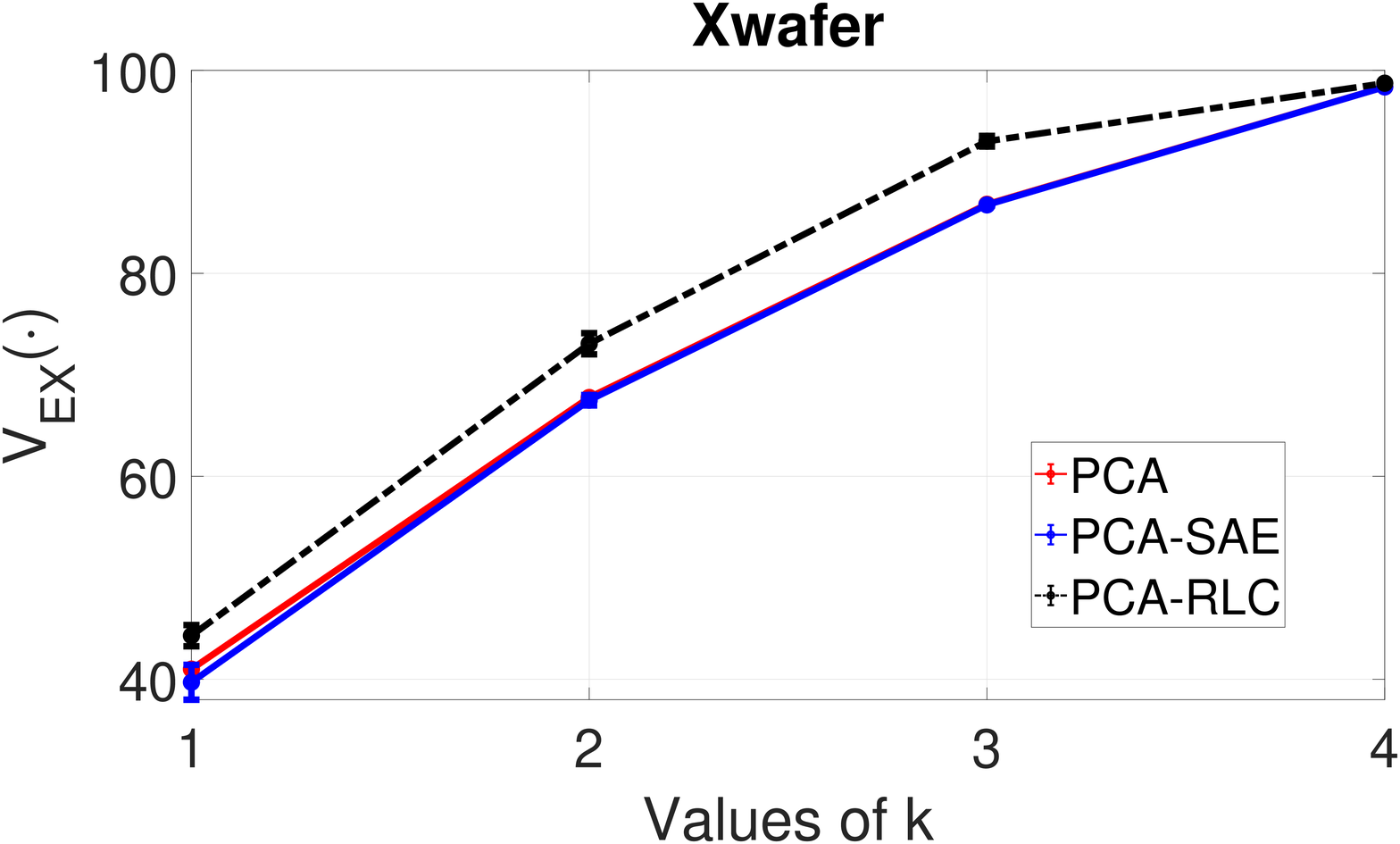} &
\includegraphics[width=0.23\textwidth]{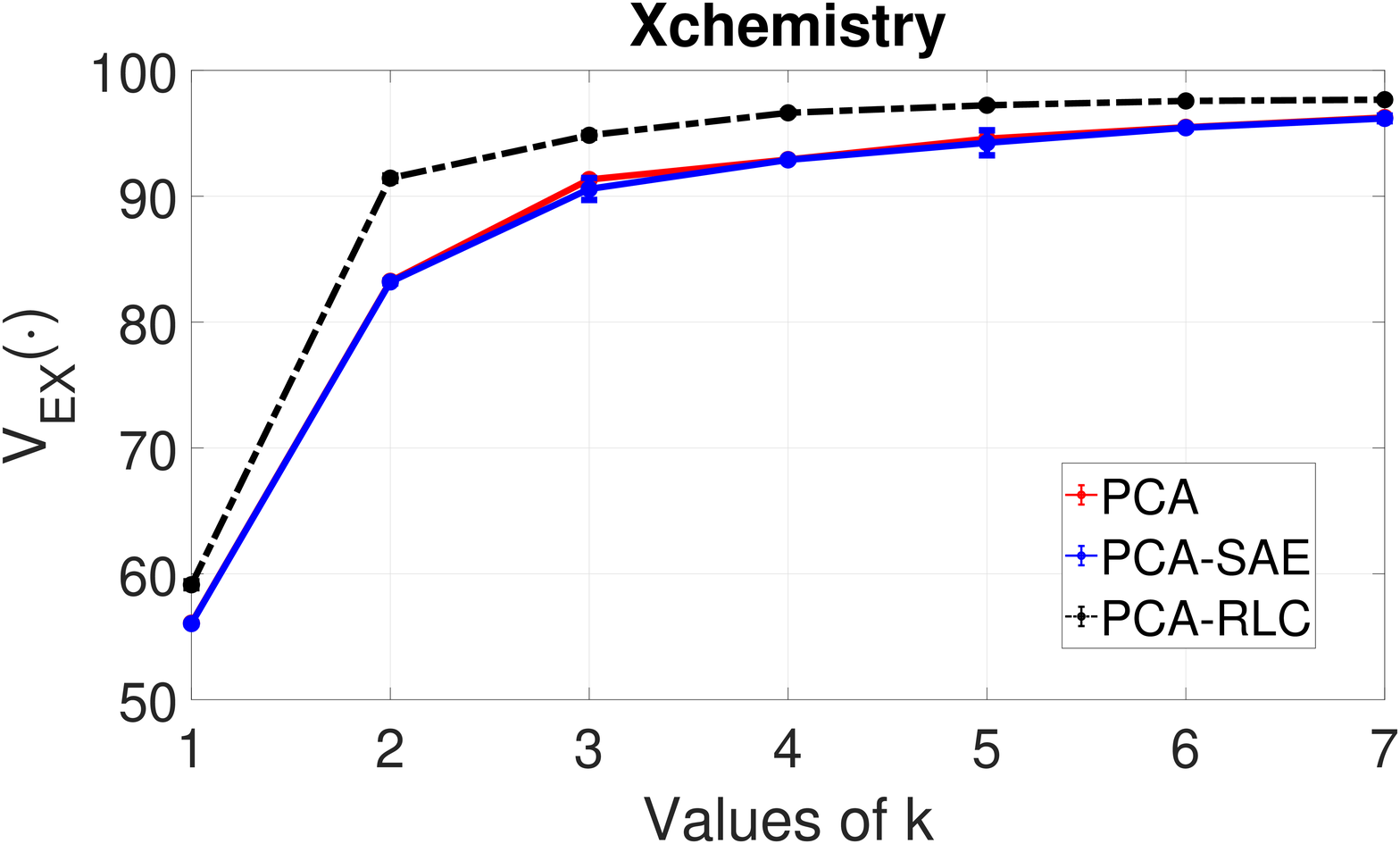} &
\includegraphics[width=0.23\textwidth]{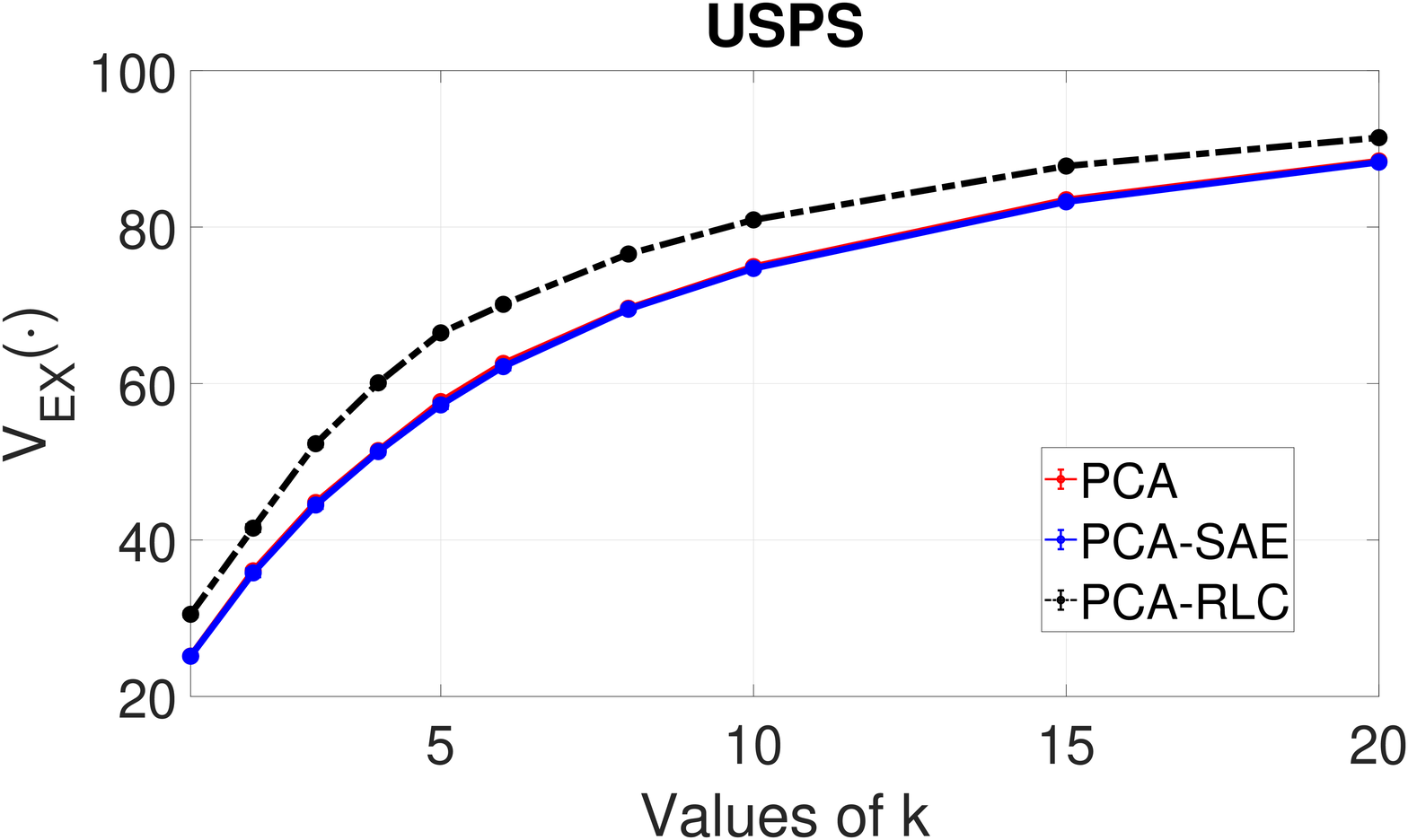} \\
\end{tabular}
\begin{tabular}{c}
\small (c) [\emph{Scenario C}] The same training/test sets used in all Monte Carlo simulations, PCA-SAE trained \\ \small \emph{without} using greedy layer-wise pre-training, and an equivalence condition (\ref{Eq:equivalenceCond}) imposed.\\[6pt] 
\end{tabular}
\caption{Variance explained be each dimensionality reduction technique as a function of $k$ for four different datasets with: (a) \emph{Scenario A}; (b) \emph{Scenario B}, and; (c) \emph{Scenario C}. The lines connect the mean values at each $k$ and the vertical bars denote the standard deviations.}
\label{fig:DimRedVEvsk}
\end{figure*}

Fig. \ref{fig:DimRedVEvsk} shows the variance explained as a function of $k$ for the four different datasets and the three training scenarios. Overall Fig. \ref{fig:DimRedVEvsk}(a) shows that PCA-SAE is the most accurate method followed by PCA-RLC and PCA. While in Xsynthetic PCA-RLC has almost equivalent performance to PCA-SAE, in the other case studies PCA-SAE significantly outperforms PCA-RLC at lower $k$ values. With the USPS dataset PCA-SAE struggles at the higher values of $k$ and is the worst performing of the three methods for $k>10$. The variance in performance over the Monte Carlo simulations is relatively small for each algorithm and value of $k$, which confirms the achievement of two positive outcomes: first, the training set contains enough information to permit us to define a model able to generalize on the test set, and second, the algorithms are robust to overfitting. 
%
Note also that the differences in variance explained between PCA-SAE and PCA-RLC is greater than the corresponding differences between FSCA-SDE and FSCA-RLC. This is simply a reflection of the fact that PCA-SAE benefits from having the additional representational power of a nonlinear encoding map whereas the other methods (including FSCA-SDE) have an encoding map that is entirely linear. 

The most striking feature of the results for \emph{Scenario B} (Fig. \ref{fig:DimRedVEvsk}(b)) is the impact on the performance of PCA-SAE of not employing greedy pre-training. This autoencoder is now the worst performing method, particularly for lower values of $k$, with both a lower mean and much higher dispersion in $V_{EX}$ for each value of $k$, highlighting the importance of performing pre-training to regularize deep neural networks. Since the training and test datasets remain the same for all simulations in this scenario, the observed dispersion is caused exclusively by the sub-optimal neural network training process. In contrast to PCA-SAE, the dispersion in $V_{EX}$ decreases for PCA-RLC relative to \emph{Scenario A} and is zero for PCA. PCA-SAE shows contradictory behaviour for higher values of $k$ in the USPS case study as it achieves superior results to the other methods, whereas the opposite was the case for \emph{Scenario A}.

Overall PCA-RLC yields stable training performance at each $k$ for all case studies and does so without the use of any pre-training. This can largely be attributed to the lower complexity neural network models that arise with the RLC topology compared to PCA-SAE, as detailed in Table \ref{tab:DimRedSettings}. The PCA-RLC neural network $\mathcal{N}$  has fewer neurons and only a single-hidden layer, leading to a less challenging training problem than with the deeper PCA-SAE architecture. Therefore, as a final evaluation, \emph{Scenario C} compares the performance of the two approaches when they are constrained to have the same number of parameters and layers, as expressed in the following equivalence condition (\ref{Eq:equivalenceCond}):   
\begin{equation} 
\theta_{SAE} = kk_{lin} + k + kk_{lin} + k_{lin} = 2kk_{lin} + k + k_{lin} 
\label{Eq:PCA-SAE}
\end{equation}
\begin{equation}
\theta_{RLC} = kh + h + h\bar{k} + \bar{k} = hk_{lin} + h + k_{lin} - k  
\label{Eq:PCA-RLC}
\end{equation}
\begin{equation}
\theta_{SAE} = \theta_{RLC}  \Rightarrow h = 2k  
\label{Eq:equivalenceCond}
\end{equation}       
Equation (\ref{Eq:PCA-SAE}) defines the number of parameters of a single-hidden layer $\mathcal{N}$ PCA-SAE architecture, while equation (\ref{Eq:PCA-RLC}) is the equivalent expression for a PCA-RLC with $h$ hidden neurons. In both topologies the neural network hidden layers are implemented with hyperbolic tangent activation functions, while the output layers are linear. 

The results for \emph{Scenario C} are presented in Fig. \ref{fig:DimRedVEvsk}(c). Since PCA-SAE is not employing a deep network, as was the case in \emph{Scenario B}, it has much more stable training performance even without pre-training, but its representational capabilities are greatly diminished with the result that it can only achieve the same level of explained variance as linear PCA.  Under these conditions PCA-RLC is substantially superior to PCA-SAE while also having much shorter training times (see Table \ref{tab:DimRed-CompTime}). 

\begin{figure*}[!t]
\vspace{5pt}
\centering
\begin{tabular}{cccc}
\includegraphics[width=0.23\textwidth]{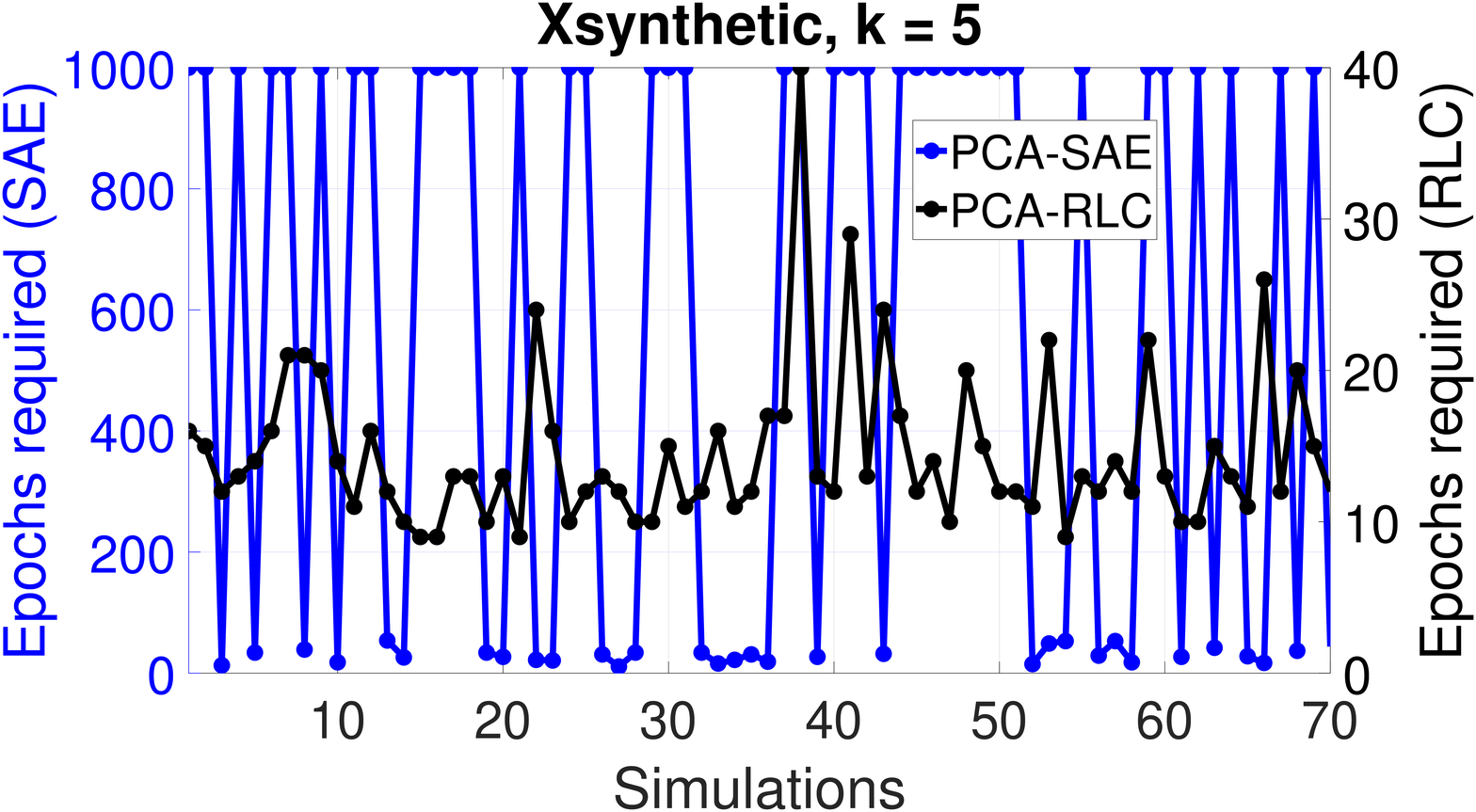} &
\includegraphics[width=0.23\textwidth]{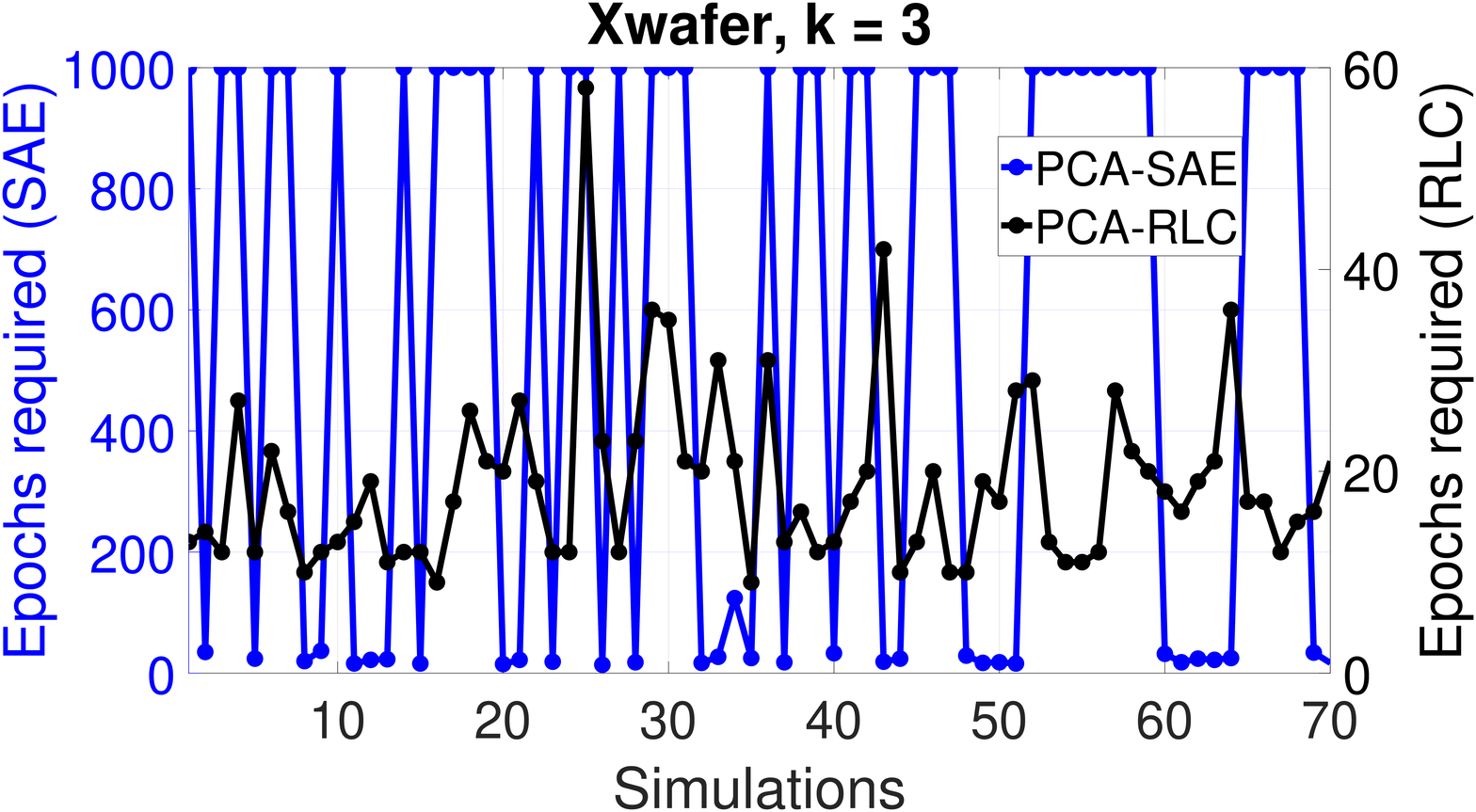} &
\includegraphics[width=0.23\textwidth]{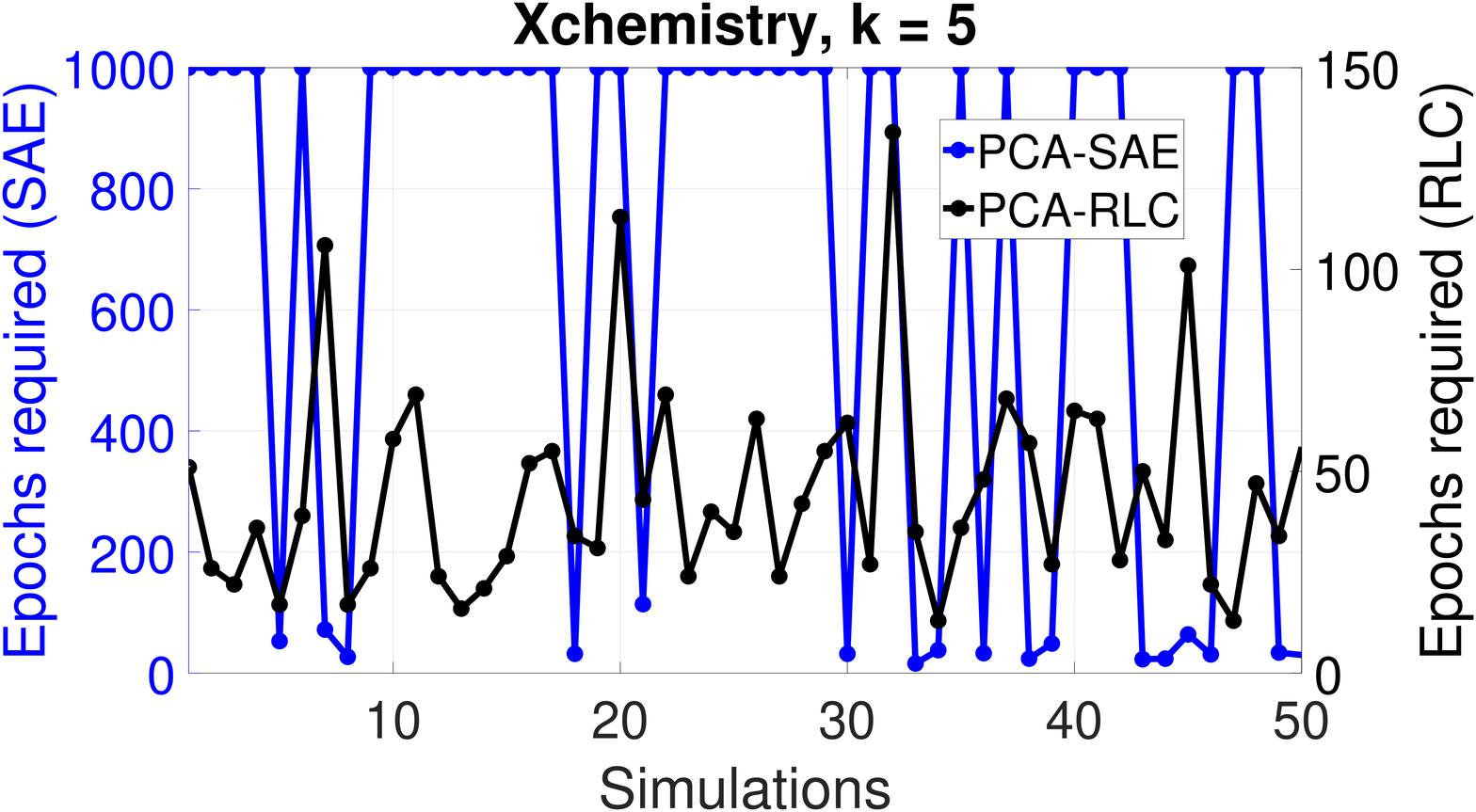} &
\includegraphics[width=0.23\textwidth]{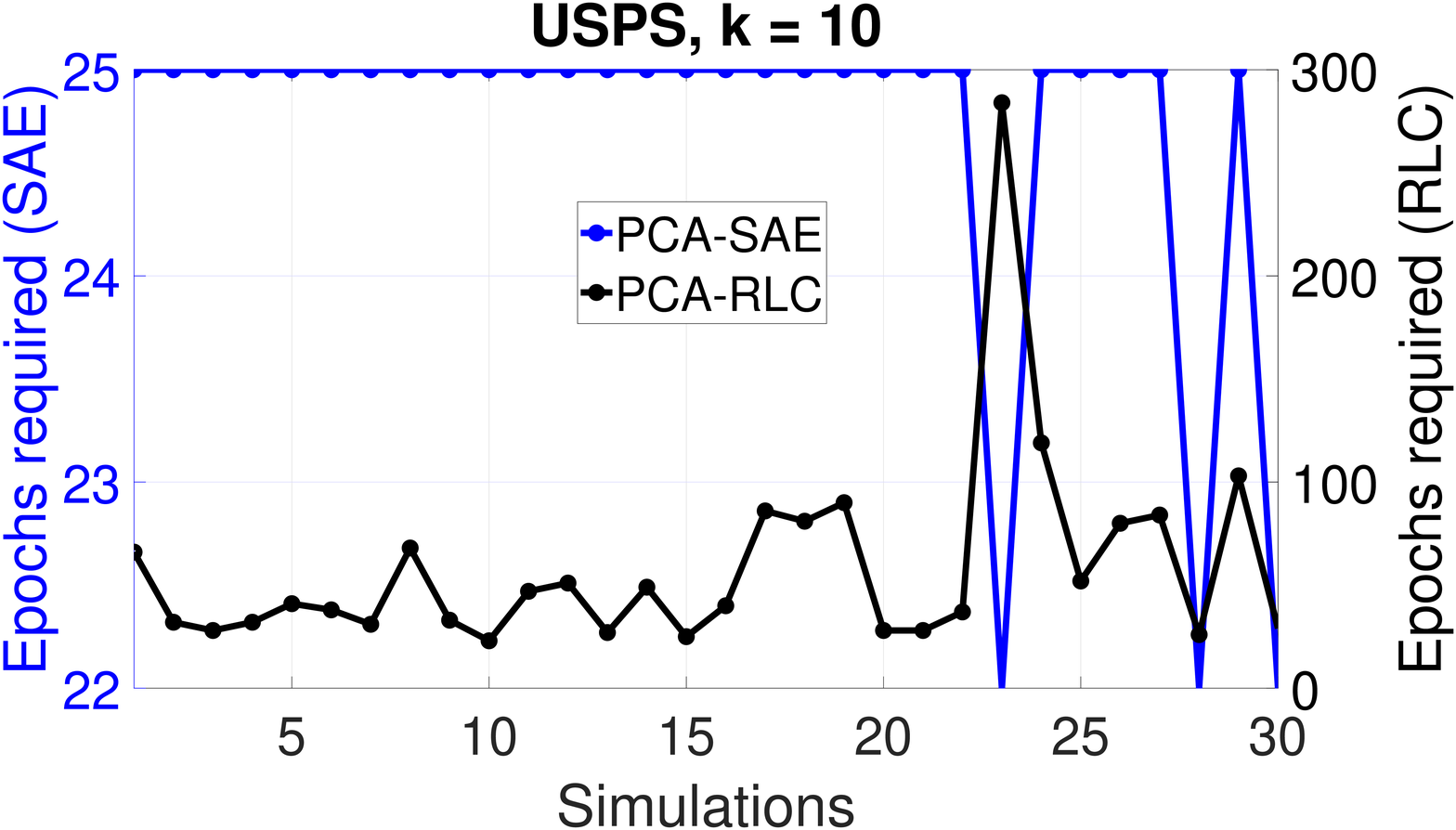}\\
\end{tabular}
\caption{Number of epochs required in each Monte Carlo simulation to train $\mathcal{N}$ in PCA-SAE and PCA-RLC for \emph{Scenario C}. Early stopping is employed for regularization and the maximum number of epochs allowed is as specified in Table \ref{tab:DimRedSettings}.}
\label{fig:dimRedEpochsPerSimulation}
\end{figure*}

\begin{figure*}[!t]
\vspace{5pt}
\centering
\begin{tabular}{cccc}
\includegraphics[width=0.23\textwidth]{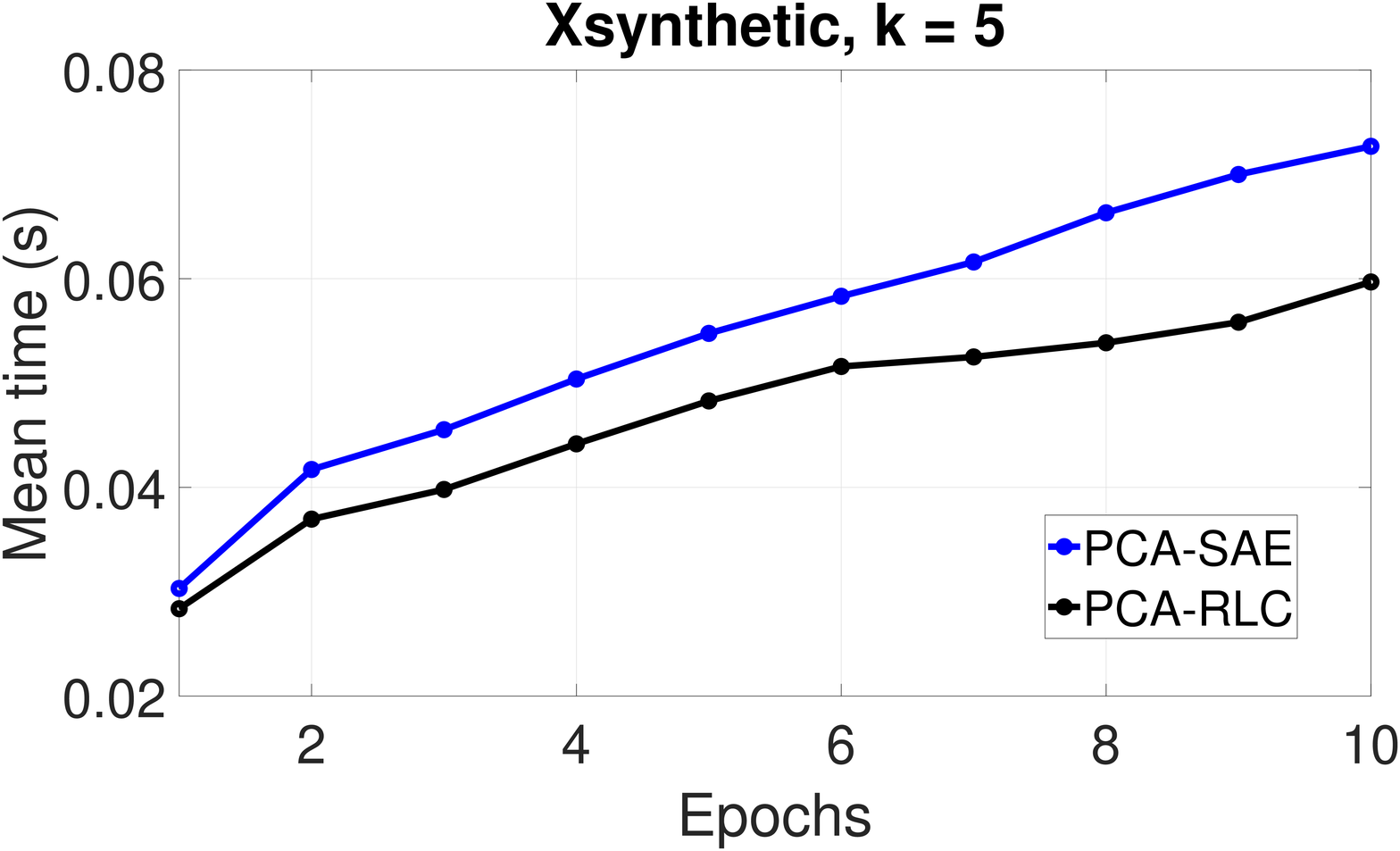} &
\includegraphics[width=0.23\textwidth]{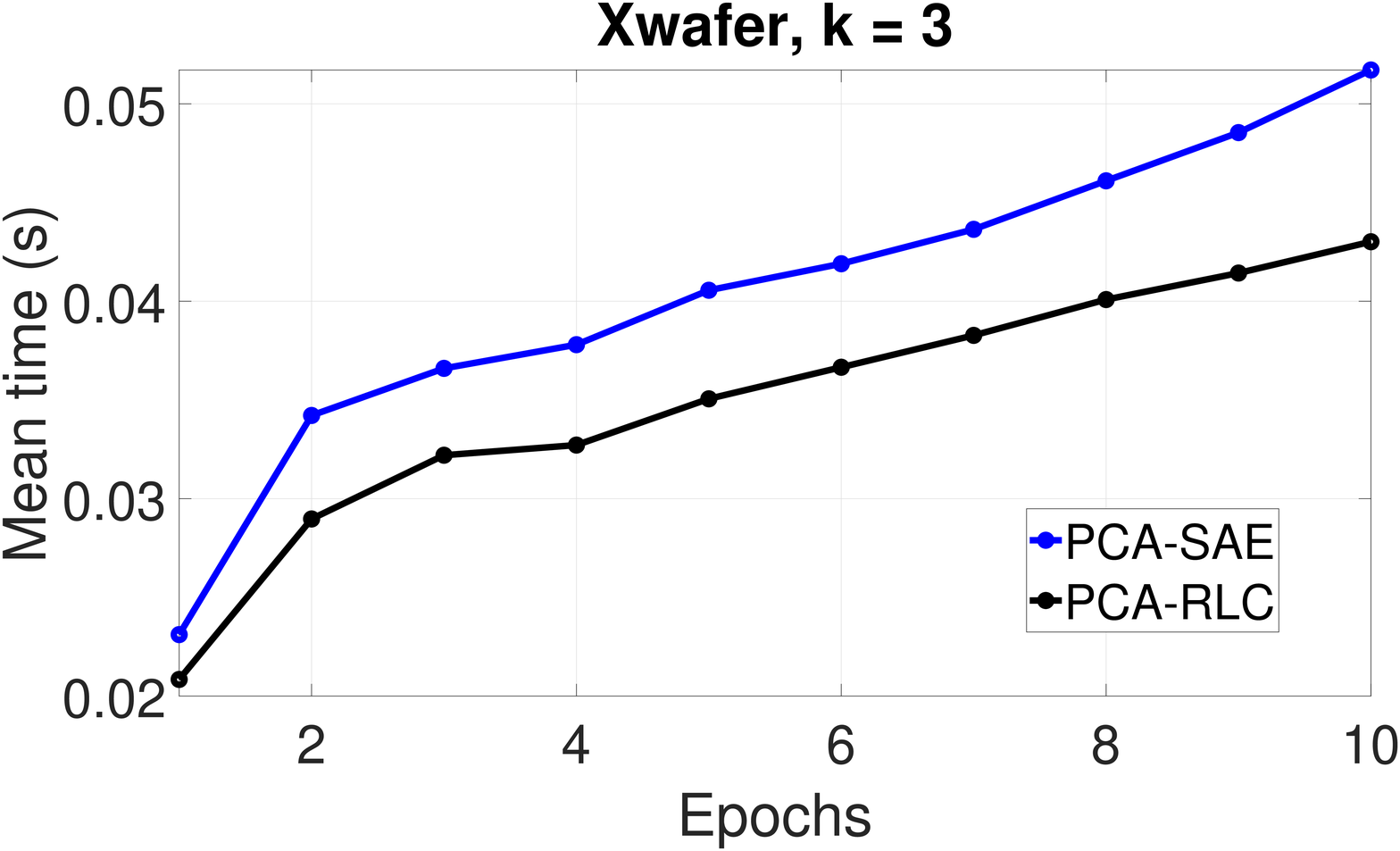} &
\includegraphics[width=0.23\textwidth]{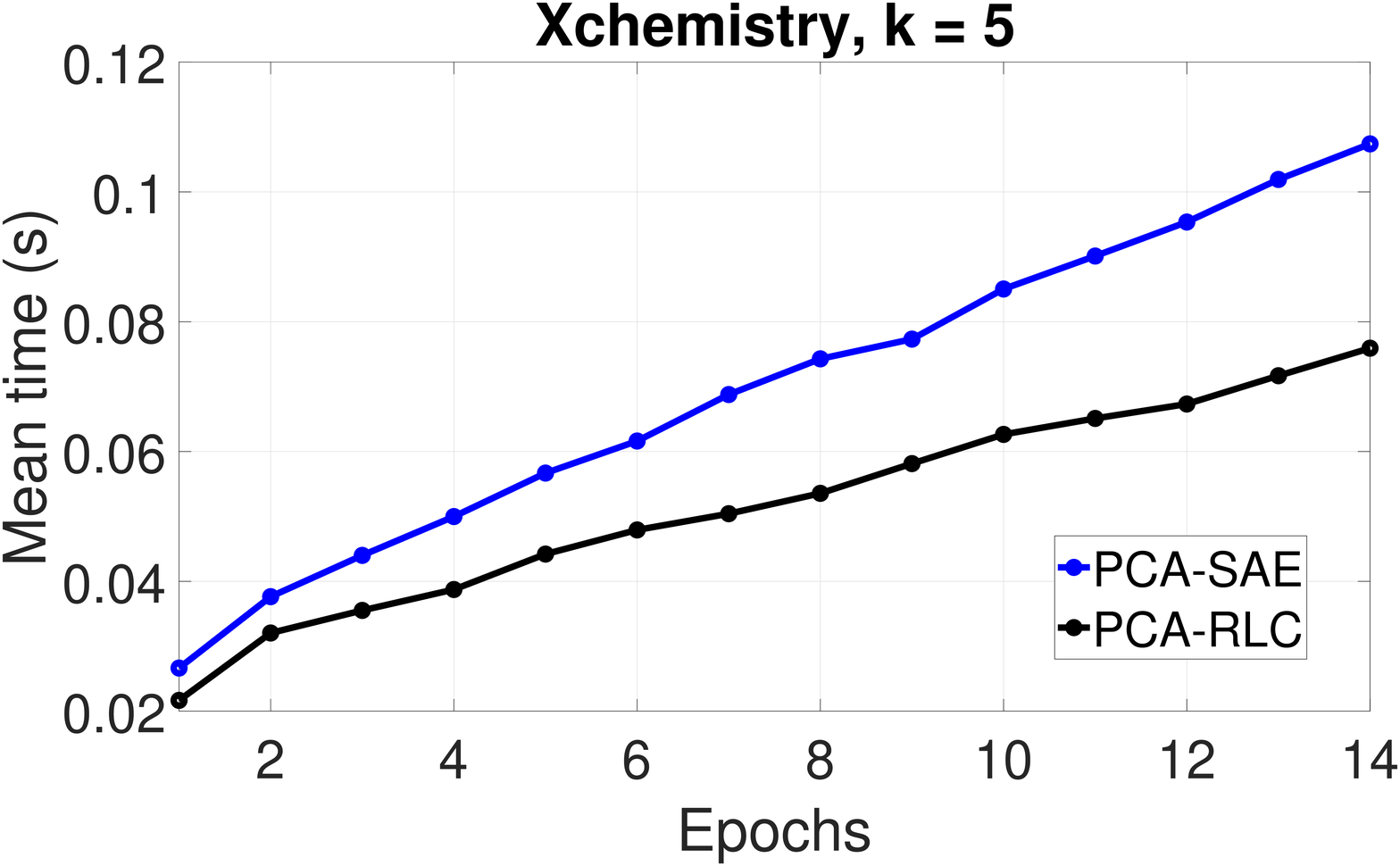} &
\includegraphics[width=0.23\textwidth]{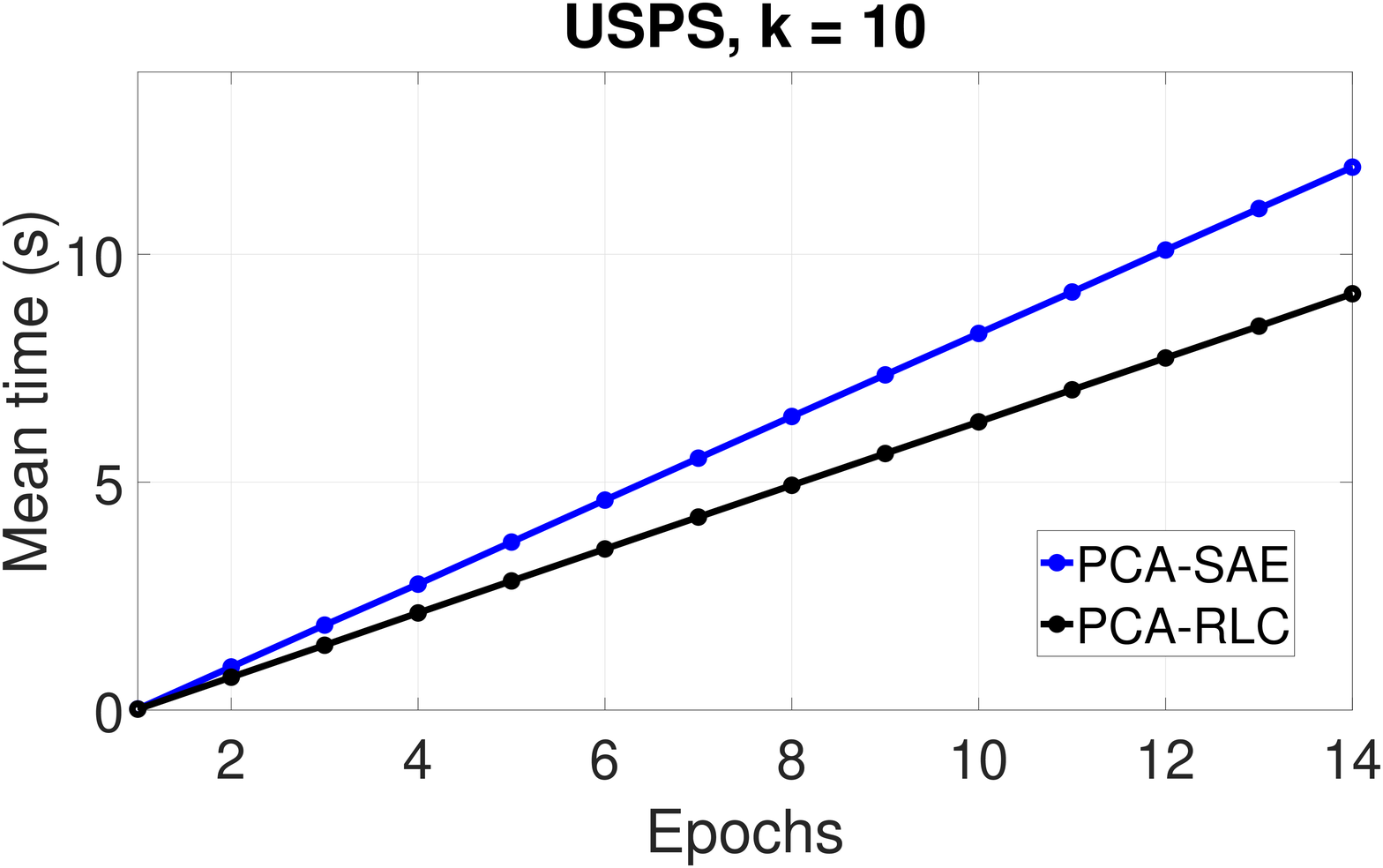}\\
\end{tabular}
\caption{The mean computation times (averaged over the Monte Carlo simulations) for the first few epochs of the training process for PCA-SAE and PCA-RLC when operating in accordance with \emph{Scenario C}.}
\label{fig:dimRedTimePerEpoch}
\end{figure*}

To gain some insight into why training times are much shorter for PCA-RLC than for PCA-SAE in \emph{Scenario C}, even though both networks have the same number of weights, the number of epochs required to train each network in each simulation and the mean computational time over the first few epochs are plotted in Figures \ref{fig:dimRedEpochsPerSimulation} and \ref{fig:dimRedTimePerEpoch}, respectively, for the four case studies. Two observations can be made with regard to these results: first, recalling that early stopping is employed as part of the training process, the total number of epochs required to train a PCA-RLC averages between 15 and 50 and seldom exceeds 100, whereas early stopping rarely interrupts the training of PCA-SAE which often reaches the 1000 limit on the number of epochs allowed (see Table \ref{tab:DimRedSettings}); and second, performing a single epoch of PCA-RLC is on average 30\% faster than one epoch of PCA-SAE. This suggests that the RLC framework leads to inherently better conditioned and therefore easier to solve training problems than those that arise with the standard SAE architecture. Note that the different pattern observed with PCA-SAE on USPS in Fig. \ref{fig:dimRedEpochsPerSimulation} is due to the limit of 25 epochs imposed on PCA-SAE training for this case study (instead of 1000). This was necessary due to the prohibitive training times for the SAE with this size of dataset.

The PCA-RLC and PCA-SAE autoencoder training times for \emph{Scenario A} and  \emph{Scenario C}, expressed as fraction of the computation time for PCA, are given in Table \ref{tab:DimRed-CompTime}. Overall, PCA-SAE takes between 30 and 400 times longer to train than PCA-RLC for \emph{Scenario A}, and between 5 and 25 times longer for \emph{Scenario C}, with the differential increasing with larger values of $k$.


\begin{table*}[!t]
	\centering
	\caption{Mean(Standard Deviation) of the Computation Time for PCA-SAE and PCA-RLC Expressed as a Fraction\\ of the Computation Time for PCA for Different Values of $k$ in \emph{Scenario A} and \emph{Scenario C}. \\The Fastest Algorithm for Each Dataset and Each $k$ is Highlighted in Bold.}
	\begin{tabular}{c|c@{\hskip 0.1in}c@{\hskip 0.1in}c|c@{\hskip 0.1in}c@{\hskip 0.1in}c|c@{\hskip 0.1in}c@{\hskip 0.1in}c|c@{\hskip 0.1in}c@{\hskip 0.1in}c}
		& \multicolumn{3}{|c|}{Xsynthetic} &\multicolumn{3}{|c|}{Xwafer} &\multicolumn{3}{|c|}{Xchemistry} & \multicolumn{3}{|c}{USPS}\\
		& $k$ &\emph{Scenario A} & \emph{Scenario C} & $k$ & \emph{Scenario A} & \emph{Scenario C} & $k$ & \emph{Scenario A} & \emph{Scenario C} & $k$ & \emph{Scenario A} & \emph{Scenario C}\\ 
		\hline
		\multirow{3}{*}{\pbox{20cm}{PCA-SAE \\ $\div10^3$}} & 3 & 4.5(1.2) & 0.9(1.0) & 2 & 2.0(0.5) & 0.3(0.6) & 3 & 3.6(2.5) & 0.4(0.7) & 5 & 11.4(2.2) & 0.2(0.1)\\
		& 5 & 3.5(2.2) & 0.5(1.1) & 3 & 1.7(0.5) & 0.3(0.7) & 5 & 4.0(15.6) & 0.5(3.7) & 10 & 6.7(1.2) & 0.4(0.2)\\
		& 7 & 2.9(1.0) & 0.5(0.5) & 4 & 2.0(1.2) & 0.5(0.5) & 7 & 3.4(11.5) & 0.3(1.7) & 20 & 3.9(0.5) & 0.7(0.02)\\
		\hline
		\multirow{3}{*}{PCA-RLC}  & 3 & 120.6(172.5) & $\bm{87.2(61.7)}$ & 2 & 67.9(26.8) & $\bm{52.6(24.2)}$ & 3 & $\bm{46.0(25.1)}$ & 59.1(70.0) & 5 & $\bm{110.8(625.5)}$ & 300.9(905.7)\\
		& 5 & 50.3(39.1) & $\bm{35.9(13.0)}$ & 3 & 53.7(30.9) & $\bm{40.6(13.6)}$ & 5 & $\bm{34.0(84.3)}$ & 44.1(141.8) & 10 & $\bm{53.6(945.3)}$ & 418.3(1764.4)\\
		& 7 & 27.1(10.0) & $\bm{20.4(2.6)}$ & 4 & 49.6(24.8) & $\bm{37.0(8.5)}$ & 7 & $\bm{20.0(44.8)}$ & 21.2(31.0) & 20 & $\bm{9.9(46.4)}$ & 452.7(1573.8)\\
		\hline
	\end{tabular}
	\label{tab:DimRed-CompTime}
\end{table*}


\section{Conclusions}
\label{sec:Concl}
Recovery of Linear Components (RLC) has been proposed as a novel methodology for unsupervised dimensionality reduction and variable selection, whereby starting with a set of $k_{lin}$ ordered linear components, a neural network is used to recover the full set of components from a subset of the first $k$ components. Crucially, by employing a linear encoding block and a nonlinear decoding block, RLC provides a compromise between fully nonlinear autoencoders and linear techniques such as PCA and FSCA, delivering improved representational capabilities over linear methods, while requiring significantly lower complexity autoencoder designs than conventional SAEs.  

The FSCA and PCA implementations of RLC (PCA-RLC and FSCA-RLC), representative of variable selection and dimensionality reduction formulations, respectively, have been presented and evaluated on a synthetic dataset and several real world case studies. The results confirm that RLC is able to achieve significantly higher levels of variance explained than linear techniques when high levels of compression are required (i.e. the number of components $k$ is relatively small). It is also robust to overfitting and substantially faster to train than SAEs that can achieve similar or better performance. In particular, for the specific case of the semiconductor manufacturing wafer measurement plan optimisation problem that motivated this work, FSCA-RLC is able to achieve state-of-the-art performance with a modest increase in computational complexity, and is two orders of magnitude faster to train than the equivalently performing FSCA-SDE.


\section*{Acknowledgments}
The first author gratefully acknowledges Irish Manufacturing Research (IMR) for the financial support provided for this work.

\ifCLASSOPTIONcaptionsoff
  \newpage
\fi



\bibliographystyle{IEEEtran}
\bibliography{References}
\end{document}